%% file: main.tex
\documentclass[10pt,twocolumn,letterpaper]{article}

\usepackage[pagenumbers]{cvpr} 

\input{preamble}
\definecolor{cvprblue}{rgb}{0.21,0.49,0.74}
\usepackage[pagebackref,breaklinks,colorlinks,allcolors=cvprblue]{hyperref}

\usepackage[accsupp]{axessibility}  

\usepackage{stfloats}
\usepackage[T1]{fontenc}
\usepackage{tabularx}

\definecolor{mygreen}{rgb}{0.267, 0.596, 0.361}

\definecolor{tablegreen}{HTML}{E6FFE6}
\definecolor{tableblue}{HTML}{E6F0FC}
\definecolor{tablegray}{HTML}{F1F4F7}
\definecolor{single_image_green}{RGB}
{61,123,178}
\definecolor{multi_image_orange}{RGB}
{188,187,214}
\newcommand{\colorcell}{\cellcolor{tableblue}}

\newcommand{\categoryentry}[3]{%
    \tikz[baseline=-0.25em]{%
        \node[
            fill=#1,          
            text=white,       
            rectangle,
            rounded corners=0pt,  
            minimum height=.2em,
            inner sep=1pt,
            outer sep=0pt
        ] {#2 (#3K)};
    }%
}

\newcommand{\taskentry}[4]{%
    \tikz[baseline=-0.25em]{
    \node[
    fill=#1,
    rectangle,
    minimum width=0.5em,
    minimum height=0.5em,
    inner sep=0pt,
    outer sep=0pt,
    ](box){};
    } #2 #4 (#3K) %
}

\newcommand{\subtaskentry}[3]{%
    \tikz[baseline=-0.25em]{%
        \node[
            fill=none,
            rectangle,
            minimum width=0.25em,
            minimum height=1em,
            inner sep=0pt,
            outer sep=0pt,
        ] (box) {};
        \draw[-] (box.north east) -- ++(0,-0.4em) -- ++(.9em,0);
        \node[anchor=west, inner sep=0pt, outer sep=0pt] (text) at (1.25em,0) {#1 #3}; 
    }
}

\usepackage[most]{tcolorbox}
\definecolor{takeawaybg}{RGB}{255, 246, 236}
\newtcolorbox{takeawaybox}{
  colback=takeawaybg,    
  colframe=takeawaybg,   
  boxrule=0pt,           
  arc=0pt,               
  left=6pt, right=6pt, top=4pt, bottom=4pt, 
  enhanced,              
  boxsep=0pt
}
\newcommand{\takeaway}[1]{%
  \begin{takeawaybox}
    #1
  \end{takeawaybox}
}

\definecolor{gold}{RGB}
{255,215,0}
\definecolor{starpink}{RGB}
{255,105,180} 
\definecolor{perceptionblue}{RGB}{57,139,142}
\definecolor{reasoningpink}{HTML}{CB4564}

\definecolor{backgroundpurple}{HTML}{faeef8}

\newtcolorbox{hypothesisbox}{
  colback=backgroundpurple,    
  colframe=backgroundpurple,   
  boxrule=0pt,           
  arc=0pt,               
  left=6pt, right=6pt, top=4pt, bottom=4pt, 
  enhanced,              
  boxsep=0pt
}

\newcommand{\hypothesis}[1]{%
  \begin{hypothesisbox}
    #1
  \end{hypothesisbox}
}

\newtcolorbox{promptbox}[3][]{
    enhanced,
    breakable,
    colback=white,
    colframe=#3,
    arc=6pt,
    boxrule=1pt,
    width=\linewidth,
    left=8pt, right=8pt, top=8pt, bottom=6pt,
    detach title,
    overlay={
        \node[
            anchor=south west,
            draw=#3,
            fill=#3,
            rounded corners=3pt,
            inner sep=4pt,
            text=white,
            font=\small\bfseries
        ] at ([xshift=6pt, yshift=-6pt]frame.north west)
        {#2};
    },
    #1
}

\newtcolorbox{promptboxcompact}[3][]{
    enhanced,
    breakable,
    colback=white,
    colframe=#3,
    arc=6pt,
    boxrule=0.75pt,
    width=\linewidth,
    left=6pt, right=6pt, top=10pt, bottom=4pt,
    detach title,
    before upper={\small},
    overlay={
        \node[
            anchor=south west,
            draw=#3,
            fill=#3,
            rounded corners=3pt,
            inner sep=3pt,
            text=white,
            font=\footnotesize\bfseries
        ] at ([xshift=6pt, yshift=-6pt]frame.north west)
        {#2};
    },
    #1
}

\newcommand{\bftable}{\fontseries{b}\selectfont}
\usepackage{multirow}
\usepackage{xspace}
\usepackage{colortbl}       
\usepackage{makecell}
\newcommand{\rotatecol}[1]{\rotatebox{90}{#1}}
\newcommand{\colpad}{\hspace{6pt}}

\def\paperID{XXXXX} 
\def\confName{CVPR}
\def\confYear{2026}

\title{\textit{Downscaling Intelligence}: Exploring Perception and Reasoning Bottlenecks  \\ in Small Multimodal Models}

\author{
Mark Endo, Serena Yeung-Levy\\
Stanford University\\
\footnotesize \bfseries \url{https://web.stanford.edu/~markendo/projects/downscaling_intelligence}
}

\begin{document}
\maketitle
\input{sec/0_abstract}    
\input{sec/1_intro}
\input{sec/2_related_work}
\input{sec/3_downscaling_exploration}
\input{sec/4_extract_think}
\input{sec/5_conclusion}
\input{sec/6_acknowledgements}
{
    \small
    \bibliographystyle{ieeenat_fullname}
    \bibliography{main}
}

\input{sec/X_suppl}

\end{document}

%% file: sec/0_abstract.tex
\begin{abstract}

Scaling up multimodal models has enabled remarkable advances in visual understanding and reasoning, but practical demands call for smaller, efficient systems. In this work, we conduct a principled analysis of downscaling intelligence in multimodal models, examining how reduced large language model (LLM) capacity affects multimodal capabilities. Our initial findings reveal an interesting trend: LLM downscaling disproportionately affects visual capabilities, rather than abilities inherited from the LLM. We then examine whether this drop mainly reflects the expected decline in visual reasoning or a more fundamental loss of perceptual abilities. Isolating the effect of LLM downscaling on perception, we find performance still drops sharply, often matching or exceeding the impact on reasoning. To address this bottleneck, we introduce visual extraction tuning, which explicitly trains the model to extract instruction-relevant visual details consistently across tasks. With these extracted visual details, we then apply step-by-step reasoning to generate answers. Together, these components form our \textsc{Extract+Think} approach, setting a new standard for efficiency and performance in this space.

\end{abstract}

%% file: sec/1_intro.tex
\section{Introduction}
\label{sec:intro}

\begin{figure*}[t!]
\centering\includegraphics[width=1\linewidth]{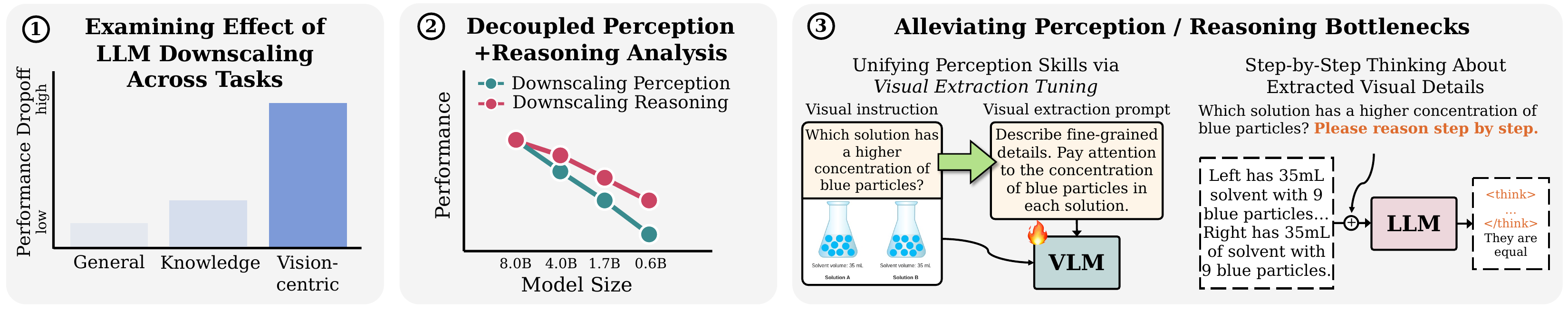}
    \caption{\textbf{Overview.} (\textit{1}) We first analyze how downscaling language model size affects multimodal performance, finding that tasks which rely more heavily on the base LLM (e.g., general or knowledge tasks) are largely unaffected, whereas visually-demanding tasks show a disproportionate drop. (\textit{2}) To uncover the mechanisms underlying the deteriorating visual capabilities under LLM downscaling, we perform a decoupled analysis of perception and reasoning, revealing that perception (alongside reasoning) is a critical bottleneck in small multimodal models. (\textit{3}) To address these limitations, we present a two-stage perception–reasoning framework, featuring \textit{visual extraction tuning}--which trains the model to extract instruction-relevant visual details consistently across tasks--coupled with step-by-step reasoning about the extracted visual details.
    }
   \label{fig:pull_figure}
\end{figure*}

Multimodal large language models (MLLMs) have become a dominant area of research in artificial intelligence, with large-scale systems demonstrating remarkable capabilities across areas spanning visual understanding and reasoning \cite{comanici2025gemini, hurst2024gpt}. Because of their impact and wide-ranging applications, increasing work has focused on understanding the scaling laws of these methods, investigating how increasing parameters and training data enhances their capabilities \cite{aghajanyan2023scaling, shukor2025scaling, tian2025navil}. However, there exists a widespread demand for smaller, efficient models suitable for on-device applications. While this need for compact architectures has spurred the development of many small models \cite{korrapati2025moondream, abdin2024phi3technicalreporthighly, marafioti2025smolvlm, lu2024deepseek}, the consequences of \textit{downscaling intelligence} remain poorly understood. Namely, when smaller language models serve as the backbone of a multimodal system, which capabilities degrade most, and \textit{why}?

In this work, we systematically investigate how downscaling large language model (LLM) size impacts multimodal behavior in order to (1) understand their practical limitations, (2) uncover the mechanisms behind their failures, and (3) develop targeted solutions to improve their performance (Figure~\ref{fig:pull_figure}). Starting with a controlled exploration across diverse visual instruction tuning tasks, we observe a striking pattern: tasks with the largest performance drop rely mainly on visual capabilities rather than the base LLM’s abilities. Based on this observation, we use a decoupled framework separating perception and reasoning, allowing us to assess whether loss of visual capabilities stems mainly from an expected decline in visual reasoning or also from a more fundamental ability to interpret and extract visual information. Notably, we find that isolating the impact of LLM downscaling on perception still results in severe performance drops across tasks, often matching or exceeding the drops observed when isolating reasoning.

To address the limitations of small multimodal models, we first focus on the discovered perception bottleneck. Because instruction tuning exposes the model to diverse ways of interpreting and utilizing visual information, we hypothesize that this bottleneck arises from the model needing to acquire diverse skills to extract relevant visual information. Thus, we propose \textit{visual extraction tuning}, a training paradigm in which the model explicitly learns to extract the visual details relevant to each instruction. We then enhance reasoning by applying step-by-step thinking over the extracted visual details, substantially enhancing performance without requiring any additional supervision on visual data. Our final two-stage approach, named \textsc{Extract+Think}, demonstrates extreme parameter and data efficiency. For example, our smaller variant surpasses the baseline two-stage PrismCaptioner framework \cite{qiao2024prism} across a wide range of tasks using a perception module roughly \textbf{\textit{12$\boldsymbol{\times}$}} smaller and a reasoning module \textbf{\textit{41$\boldsymbol{\times}$}} smaller. Even when training from scratch utilizing visual extraction tuning, our approach improves over LLaVA-OneVision-0.5B \cite{li2024llavaonevision} while using \textbf{\textit{95\%}} fewer visual training samples. Together, our work offers the first systematic characterization of downscaling effects in multimodal models and introduces effective solutions to their bottlenecks, laying the groundwork for future 
advances in small-scale multimodal intelligence.

%% file: sec/2_related_work.tex
\section{Related Work}
\label{sec:related_work}

\textbf{Small MLLMs.} The development of small yet powerful vision-language models (VLMs) has been a significant focus of recent research, aiming to provide strong multimodal capabilities in resource-constrained environments. This includes models like Moondream \cite{korrapati2025moondream},  Phi-3-Vision 
\cite{abdin2024phi3technicalreporthighly}, SmolVLM \cite{marafioti2025smolvlm}, and MiniCPM \cite{hu2024minicpm}, as well as compact variants of Gemma 3 \cite{team2025gemma3}, DeepSeek-VL \cite{lu2024deepseek}, Qwen-VL series \cite{wang2024qwen2vl, bai2025qwen2_5vl, qwen3vl}, LLaVA-OneVision \cite{li2024llavaonevision}, and InternVL \cite{chen2024internvl2}. While these models demonstrate impressive general capabilities, their failure modes--especially those concerning visual capabilities, and in particular perception--remain poorly understood. Findings across prior works are inconsistent: some studies suggest that scaling LLM size has little effect on perception \cite{qiao2024prism, li2024llavaonevision}, while others find that perception-heavy tasks such as OCR and Chart VQA are highly sensitive to model size \cite{han2025learning}. These discrepancies highlight the need for an in-depth analysis, which we undertake to examine how downscaling LLMs affects visual capabilities and the mechanisms behind their failures.

\textbf{Failures of MLLMs.} A number of works have revealed shortcomings of state-of-the-art MLLMs on perceptual and visual reasoning tasks. For example, \cite{fu2024blink} discovers that even the best-performing multimodal models perform near-randomly on perceptual tasks that humans can solve quickly. For visual spatial planning, \cite{wu2024vsp} identifies fundamental deficiencies in the models’ visual perception and reasoning abilities. Many works demonstrate that VLMs often struggle with visual reasoning puzzles that require strong pattern recognition and abstract reasoning \cite{chia2024puzzlevqa, malkinski2024bongard, wust2024bongard}. Examining why VLMs fail on visual tasks, studies often find that visual information from the vision encoder is inadequately utilized by the language model \cite{fu2025hidden, zhang2024visually, liu2025perception}, attributing failures to limited exposure to relevant visual data and mitigating this with more representative training data. However, these works often focus on much bigger and more powerful models, leaving the failures from LLM downscaling largely unexplored.

%% file: sec/3_downscaling_exploration.tex
\section{LLM Downscaling Exploration}
\label{sec:preliminary_exploration}

\begin{figure*}[b!]
    \centering
    \includegraphics[width=1\linewidth]{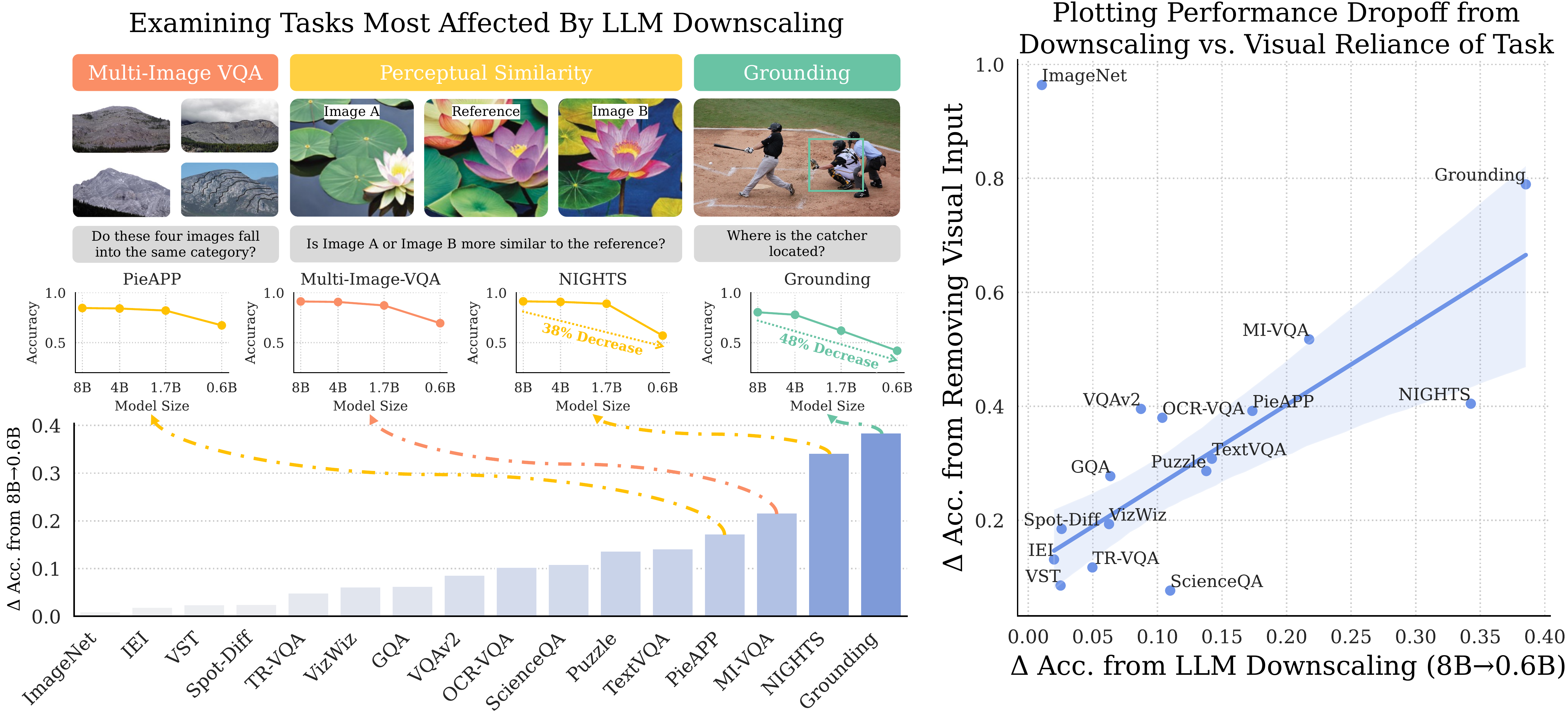}
    \caption{\textbf{LLM downscaling exploration.} (\textit{Left}) \textbf{Performance dropoff from LLM downscaling most notable for visually demanding tasks.} Tasks like \textcolor[HTML]{69C3A4}{Grounding} and \textcolor[HTML]{FFC102}{Perceptual Similarity} (e.g., NIGHTS and PieAPP) which primarily focus on visual processing are most affected by LLM downscaling, rather than tasks which rely heavily on the base LLM (such as ScienceQA evaluating knowledge or GQA assessing general abilities). (\textit{Right}) \textbf{The more a task's performance declines under LLM downscaling, the greater it depends on visual information}. As the impact of LLM downscaling increases (8B $\rightarrow$ 0.6B), so does the task’s reliance on visual information (measured by performance difference with and without visual input). IEI=Image Edit Instruction, VST=Visual Story Telling, Spot-Diff=Spot the Difference, TR-VQA=Text-Rich VQA, MI-VQA=Multi-Image-VQA. Full plots for all datasets are provided in the supplemental material.}
    \label{fig:per_task_results}
\end{figure*}

In the first part of this work, we conduct a controlled study to examine how reducing language model size impacts multimodal task performance, aiming to understand the limitations of small models as general visual assistants and the causes of their failures. After covering model and setup preliminaries (\S\ref{sec:preliminaries}), we present our results on how downscaling model size impacts performance across various tasks (\S\ref{sec:downscaling_results}). We find that the tasks most affected by downscaling language model size are \textbf{not} those that heavily rely on the base LLM, but rather those emphasizing visual processing.
Next, we investigate whether the decline in performance under LLM downscaling stems primarily from weakened visual reasoning or if it also reflects a more fundamental impairment to perception (\S\ref{sec:who_blinks_first}). When isolating the effect of LLM downscaling on perception, we find that performance still drops sharply--often matching or exceeding the decline observed when isolating its effect on reasoning--indicating that a central limitation of small multimodal models arises from a degradation in their ability to recognize and understand visual information.

\subsection{Preliminaries}
\label{sec:preliminaries}

In this work, we focus on the popular multimodal LLM approach of taking a language model trained on a broad corpora of text as the foundation, integrating a pre-trained vision encoder with a simple projector to connect the visual representations to the LLM token space, and training the combined system with visual instruction tuning data. We go over details about each component of our setup below. 

\textbf{Architecture.} Our architectural decisions are guided by the principle of using well-established and widely validated design choices, ensuring that the findings are broadly applicable and impactful for future work. Hence, we use Qwen3 series (8B, 4B, 1.7B, and 0.6B sizes) \cite{yang2025qwen3} for the LLM, SigLIP \cite{zhai2023siglip} as the vision encoder, and a 2-layer MLP as the connector. We use the Higher AnyRes with Bilinear Interpolation scheme from \cite{li2024llavaonevision} for visual processing.

\textbf{Data.} We use a broad range of visual instruction tuning datasets for our exploration. To enable a more controlled setting for analyzing task performance, we focus on data that includes both training sets and benchmark evaluations. Specifically, for single-image tasks we leverage \cite{chen2024coin}, and for multi-image tasks we utilize the subset of M4-Instruct data that includes evaluation benchmarks \cite{li2024llavanextinterleave}. We additionally include PieAPP \cite{prashnani2018pieapp} to ensure sufficient data for the Perceptual Similarity task. All datasets are listed in Table~\ref{tab:data}. For the connector pretraining stage, we utilize BLIP558K.

\textbf{Training Recipe.} Based on \cite{li2024llavaonevision}, after pre-training the connector for language-image alignment, we perform visual instruction tuning, fine-tuning all parameters on single-image data (574K) and then on a combination of the multi-image data (309K) and 150K randomly sampled single-image examples. We use the same batch size, learning rates for model parameters, and image resolutions as \cite{li2024llavaonevision}.

\begin{table}[t]
    \centering
    \setlength{\belowcaptionskip}{-2em}
    \begin{scriptsize}
    \begin{tabularx}{.85\linewidth}{@{}X@{}X@{}}
    \multicolumn{2}{c}{\small Visual Instruction Tuning Data} \\
    \noalign{\vskip 0em}
    \hline
    \noalign{\vskip .2em}
    \categoryentry{single_image_green}{Single-Image}{574} & \taskentry{single_image_green!100}{OCR-VQA}{165}{\cite{mishra2019ocrvqa}}\\
    \taskentry{single_image_green!50}{VQAv2}{82.8}{\cite{goyal2017vqav2}} & \taskentry{single_image_green!79}{ImageNet}{130}{\cite{deng2009imagenet}}\\
    \taskentry{single_image_green!12}{VizWiz}{20.5}{\cite{gurari2018vizwiz}} & \taskentry{single_image_green!34}{Grounding}{55.9}{}\\
    \taskentry{single_image_green!8}{ScienceQA}{12.7}{\cite{lu2022scienceqa}} & \subtaskentry{RefCOCO}{}{\cite{kazemzadeh2014refcoco}}\\
    \taskentry{single_image_green!21}{TextVQA}{34.6}{\cite{singh2019textvqa}} & \subtaskentry{RefCOCO+}{}{\cite{mao2016refcocoplusg}}\\
    \taskentry{single_image_green!44}{GQA}{72.1}{\cite{hudson2019gqa}} & \subtaskentry{RefCOCOg}{}{\cite{mao2016refcocoplusg}}\\
    \noalign{\vskip 0em}
    \hline
    \noalign{\vskip .2em}
    \categoryentry{multi_image_orange}{Multi-Image}{309} & \taskentry{multi_image_orange!31}{Text-Rich VQA}{21.3}{}\\
    \taskentry{multi_image_orange!43}{Spot the Difference}{28.9}{} & \subtaskentry{WebQA}{9.3}{\cite{chang2022webqa}}\\
    \subtaskentry{Spot-the-Diff}{10.8}{\cite{jhamtani2018spotthediff}} & \subtaskentry{TQA}{8.2}{\cite{kembhavi2017tqa}}\\
    \subtaskentry{Birds-to-Words}{14.2}{\cite{forbes2019birdstowords}} & \subtaskentry{OCR-VQA}{1.9}{\cite{mishra2019ocrvqa}}\\
    \subtaskentry{CLEVR-Change}{3.9}{\cite{park2019clevrchangept1,hosseinzadeh2021clevrchangept2}} & \subtaskentry{DocVQA}{1.9}{\cite{mathew2021docvqa}}\\
    \taskentry{multi_image_orange!100}{Image Edit Instruction}{67.7}{} & \taskentry{multi_image_orange!33}{Multi-Image-VQA}{22.4}{} \\
    \subtaskentry{HQ-Edit}{50}{\cite{hui2024hqedit}} & \subtaskentry{MIT-StateCoherence}{1.9}{\cite{isola2015mitstates}}\\
    \subtaskentry{MagicBrush}{14.2}{\cite{zhang2023magicbrush}} & \subtaskentry{MIT-PropertyCoherence}{1.9}{\cite{isola2015mitstates}} \\
    \subtaskentry{IEdit}{3.5}{\cite{tan2019iedit}} & \subtaskentry{RecipeQA-ImageCoherence}{8.7}{\cite{yagcioglu2018recipeqa}}\\
    \taskentry{multi_image_orange!100}{Visual Story Telling}{67.5}{} & \subtaskentry{VISION}{9.9}{\cite{bai2023vision}} \\
    \subtaskentry{AESOP}{6.9}{\cite{ravi2021aesop}} & \taskentry{multi_image_orange!52}{Puzzle (Raven)}{35}{\cite{zhang2019raven}}  \\
    \subtaskentry{FlintstonesSV}{22.3}{\cite{gupta2018flintstonessv}} & \taskentry{multi_image_orange!98}{Perceptual Similarity}{66.4}{} \\
    \subtaskentry{PororoSV}{12.3}{\cite{li2019pororosv}} & \subtaskentry{NIGHTS}{15.9}{\cite{fu2023dreamsim}} \\
    \subtaskentry{VIST}{26}{\cite{huang2016vist}} & \subtaskentry{PieAPP}{50.5}{\cite{prashnani2018pieapp}} \\

    \end{tabularx}
    \end{scriptsize}
    \caption{List of used visual instruction tuning data. Task colors indicate their relative proportion in the data mixture.}
    \label{tab:data}
\end{table}

\begin{figure*}[b]
    \centering
    \includegraphics[width=1\linewidth]{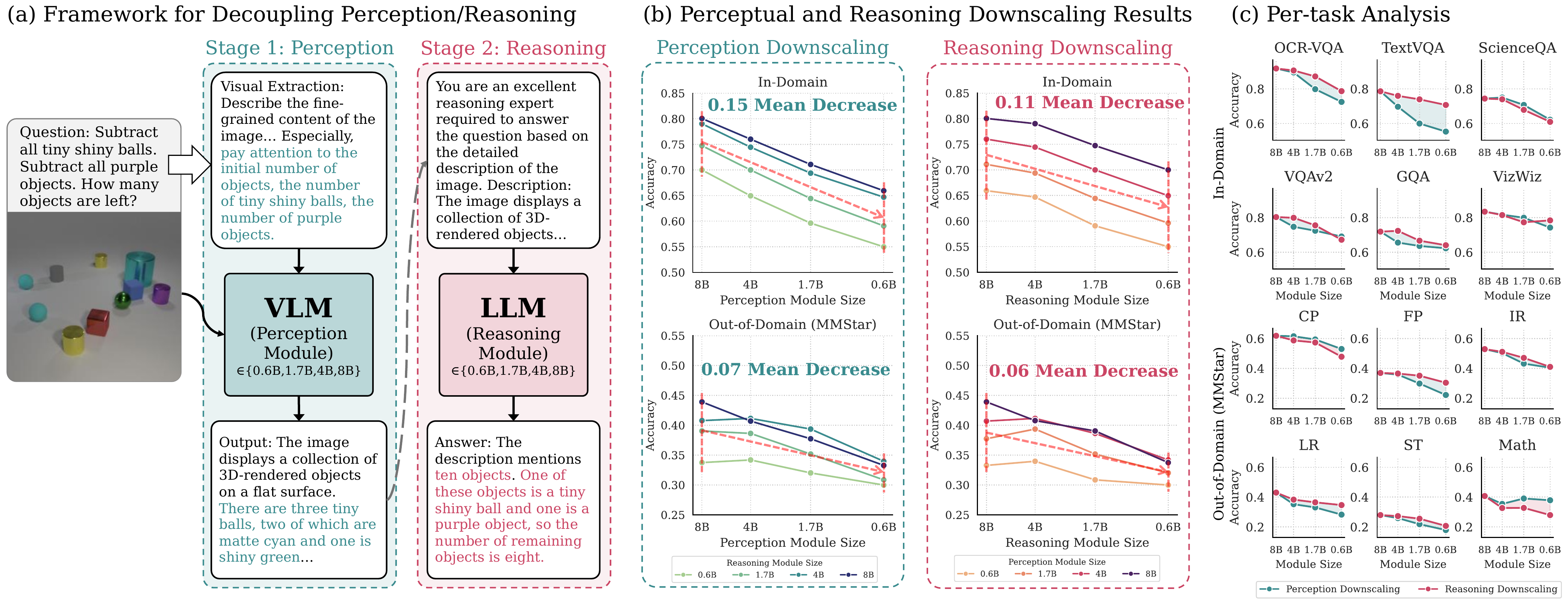}
    \caption{
    \textbf{Decoupled perception and reasoning downscaling analysis.} \textit{(a)} \textbf{Decoupled Setup.} We disentangle perceptual and reasoning abilities using a two-stage framework: the perception module (VLM) first extracts visually relevant information, then the reasoning module (LLM) generates answers based on the extracted visual information.
    \textit{(b)} \textbf{Perception and reasoning emerge as key bottlenecks under LLM downscaling.} We see that LLM downscaling of either the perception module or reasoning module largely degrades in-domain and out-of-domain task performance. \textit{(c)} \textbf{Perceptual degradation limits performance across tasks.} Even for tasks targeting visual reasoning (e.g., IR and LR), downscaling perception has an impact comparable to--or even exceeding--that of downscaling reasoning.
    In this per-task analysis, the non-downscaled module is set at 8B. CP=Coarse Perception, FP=Fine-grained Perception, IR=Instance Reasoning, LR=Logical Reasoning, ST=Science \& Technology.
    }
   \label{fig:decoupled_results}
\end{figure*}

\subsection{Analyzing impact across tasks}
\label{sec:downscaling_results}

\textbf{\textit{Tasks with largest performance drops under LLM downscaling rely heavily on visual processing, not the base LLM.}}
As shown in Figure~\ref{fig:per_task_results}(Left), most tasks exhibit modest performance decline when downscaling the language model size from 8B to 0.6B, except for a few tasks which exhibit much larger deterioration. Interestingly, rather than these tasks depending heavily on the base LLM (such as ScienceQA, which assesses knowledge, or GQA, which evaluates general abilities), they instead rely primarily on visual processing. For example, Grounding drops 48\% and NIGHTS (Perceptual Similarity) declines 38\% when downscaling from 8B to 0.6B.

\noindent \textbf{\textit{The greater the impact of LLM downscaling, the more the task relies on visual information to be solved.}}
While our analysis so far has focused on the few tasks most affected by model downscaling, here we extend the analysis to the full set of datasets. To better understand how a dataset’s sensitivity to LLM downscaling relates to how vision-centric the task is, we plot the performance difference between the 8B and 0.6B LLMs against the difference in performance with and without visual input (using the 8B LLM). As shown in Figure~\ref{fig:per_task_results}(Right), most datasets exhibit an approximately linear trend: as the impact of LLM downscaling increases, so does the task’s reliance on visual information. The exception is ImageNet, where the small model achieves very strong performance but blind performance is near zero. This likely occurs because the perception required is notably simple and the task comprises a large portion of the visual instruction tuning data.

\takeaway{\textbf{Takeaway 1:} LLM downscaling is most detrimental to vision-centric capabilities rather than base LLM abilities.}

\noindent \textit{Discussion:} While previous studies connect poor utilization of visual representations in multimodal models to limited training data \cite{zhang2024visually, fu2025hidden, liu2025perception}, we observe a distinct behavior: \textbf{\textit{even when the training mixture ensures coverage across all evaluated tasks, visually-intensive tasks deteriorate most as LLM size decreases}}. Overall, our findings suggest that in multimodal models trained using visual instruction tuning, processes related to understanding and/or reasoning about visual information are significantly impaired by downscaling the language model.

\subsection{Decoupled perception / reasoning analysis}
\label{sec:who_blinks_first}

Our findings in the previous section are intriguing, but \textbf{\textit{the reason behind the observed trend remains unclear}}. Namely, vision-centric tasks generally require two essential capabilities: \textit{perception}, the foundational ability to recognize, extract, and understand visual details, and \textit{reasoning}, the downstream ability to operate on extracted visual information to formulate answers. While our analysis showed that the visual capabilities of multimodal models degrade significantly under LLM downscaling, it did not reveal the mechanisms underlying these failures. Given that reasoning depends on model scale for textual tasks \cite{lin2025zebralogic}, we expect visual reasoning to decline under downscaling; however, the effect on the more foundational process of perception is highly uncertain and warrants further study. Thus, in this section, we perform a rigorous analysis separating the effects of LLM downscaling on perception and reasoning to better understand the causes of the observed behavior. We detail our decoupled setup for this analysis and present our results below.

\textbf{Setup.} To study perception and reasoning independently under downscaling, we apply the Prism framework \cite{qiao2024prism}, which separates these two processes. As shown in Figure~\ref{fig:decoupled_results}(a), each question is answered in two stages. In the first stage, the question is converted into a prompt to extract all visually relevant information, and both this question-specific instruction and the image are fed into a multimodal model (perception module) to obtain the important visual information. In the second stage, an LLM (reasoning module) uses the extracted visual information to reason and generate the final answer. Using this setup, we independently downscale the LLM in each module to measure how the two abilities are affected by LLM size.

In our analysis, we utilize the same multimodal models from \S\ref{sec:downscaling_results} as the perception module and their corresponding Qwen3 series models \cite{yang2025qwen3} as the reasoning module. Differing from \cite{qiao2024prism}, we convert the prompts offline using one model type (Qwen3-8B), so that the questions remain consistent across setups and the model’s ability to generate question-specific instructions does not influence our analysis of perception and reasoning. For evaluation datasets, as the reasoning module is not trained on the output distribution of the visual tasks from \S\ref{sec:preliminary_exploration}, we utilize the converted multiple-choice format of these datasets from AutoConverter \cite{zhang2025autoconverter}, which has proven to enable objective evaluation under variability in natural language responses. We exclude Grounding and ImageNet from this analysis as these are purely perceptual tasks. We additionally evaluate on the carefully curated, out-of-domain benchmark MMStar \cite{chen2024mmstar}, which assesses both perceptual and reasoning abilities.

\textbf{Results.} \noindent \textbf{\textit{LLM downscaling expectedly hinders visual reasoning.}}
As shown in Figure~\ref{fig:decoupled_results}(b), we find that downscaling the reasoning module size has a considerable impact on performance across tasks, confirming that visual reasoning is a critical bottleneck for small multimodal models.

\noindent \textbf{\textit{
LLM downscaling markedly impairs perceptual abilities, affecting a wide spectrum of tasks.}} More notably, in Figure~\ref{fig:decoupled_results}(b) we also observe that LLM downscaling of the perception module has as substantial an effect on performance, where downscaling from 8B to 0.6B causes an average accuracy drop of 0.15 for in-domain data and 0.07 for out-of-domain data. As shown in Figure~\ref{fig:decoupled_results}(c), even for tasks that target visual reasoning (such as Instance Reasoning and Logical Reasoning), downscaling the perception module has an impact on performance comparable to, or even exceeding, that of downscaling the reasoning module. This likely occurs because the foundational ability to understand visual information is a prerequisite for successfully performing downstream reasoning.

\takeaway{\textbf{Takeaway 2:} 
While LLM downscaling expectedly impairs visual reasoning, isolating its impact solely on perception still reveals severe performance degradation across a wide range of tasks, often matching or exceeding its effect on reasoning.
}

\noindent \textit{Discussion}. This section highlights an important and previously undiscovered phenomenon. The original Prism work \cite{qiao2024prism} used a relatively small LLM for the perception module (e.g., InternLM2-1.8B \cite{cai2024internlm2}) and a much larger LLM for the reasoning module (LLaMA-3-70B \cite{grattafiori2024llama} and ChatGPT)), based on the assumption that perception is far less sensitive to LLM scale than reasoning. Although reasoning is naturally expected to degrade more than perception, we find that its impact on performance is surprisingly similar to that of perceptual abilities. Thus, \textbf{\textit{perception}} \textit{(alongside reasoning)} \textbf{\textit{emerges as a central bottleneck in small multimodal models}}. Given that the visual representations are fixed across model setups, \textit{what drives this perceptual decline}?

We hypothesize that this perception bottleneck arises from a fundamental limitation of the visual instruction tuning paradigm under LLM downscaling. Namely, visual instruction tuning exposes the model to various ways of recognizing, understanding, and extracting visual information. We posit that this variability requires the model to acquire diverse skills for interpreting instructions and extracting the relevant visual information. The \textit{Quantization Model} of neural scaling laws \cite{michaud2023quantization} offers a theoretical lens: model skills can be ``quantized'' into discrete chunks (quanta), and scaling laws limit the total number a model can effectively learn from the training data. Because visual instruction tuning requires the model to learn many skills to process visual information across diverse tasks, smaller models have weaker perceptual capabilities.

\hypothesis{\textbf{Hypothesis}: LLM downscaling's effect on perception arises from the heterogeneity of how perception is learned under visual instruction tuning.}

\noindent As part of the following section, we leverage this hypothesis to guide method advancements aimed at improving the perceptual abilities of small multimodal models.

%% file: sec/4_extract_think.tex
\section{\large \textbf{\textsc{Extract+Think}}}
\label{sec:alleviating_percetual_scaling_bottleneck}

Having shown that LLM downscaling weakens both foundational perception and downstream reasoning, we conclude by proposing solutions that address these limitations and move toward a high-performing generalist small multimodal model. We focus our efforts on the two-stage framework, as it provides a modular approach to study and improve each bottleneck (though we note it adds the challenge of capturing all relevant visual information in text; see supplementary material for more details).

First, we look at improving perception of small models by streamlining the learning of perceptual skills through a new \textit{visual extraction tuning} paradigm (\S\ref{sec:improving_perception}). Next, we look at how to better utilize the extracted visual information by allowing the reasoning module to reason step-by-step (\S\ref{sec:improving_reasoning}). Together, these two components comprise our final approach \textsc{Extract+Think}, which sets a new standard in parameter- and data-efficient multimodal modeling (\S\ref{sec:final_results}), offering an effective path toward generalist small models.

\subsection{Visual extraction tuning}
\label{sec:improving_perception}

We first aim to alleviate the foundational perception bottleneck for small multimodal models.
As discussed in \S\ref{sec:who_blinks_first}, we hypothesize that the perception bottleneck on small multimodal models arises from the model needing to acquire a diverse set of skills to extract relevant visual information across a wide range of tasks. Thus, a natural approach to improve performance in downscaled LLMs is to increase the homogeneity of how visual information is extracted. In this section, we first assess captioning as a baseline method to achieve this, and then propose a new training paradigm, \textit{visual extraction tuning}, which demonstrates strong abilities in enhancing perception in small multimodal models.

\begin{figure}[t]
    \centering
    \includegraphics[width=1\linewidth]{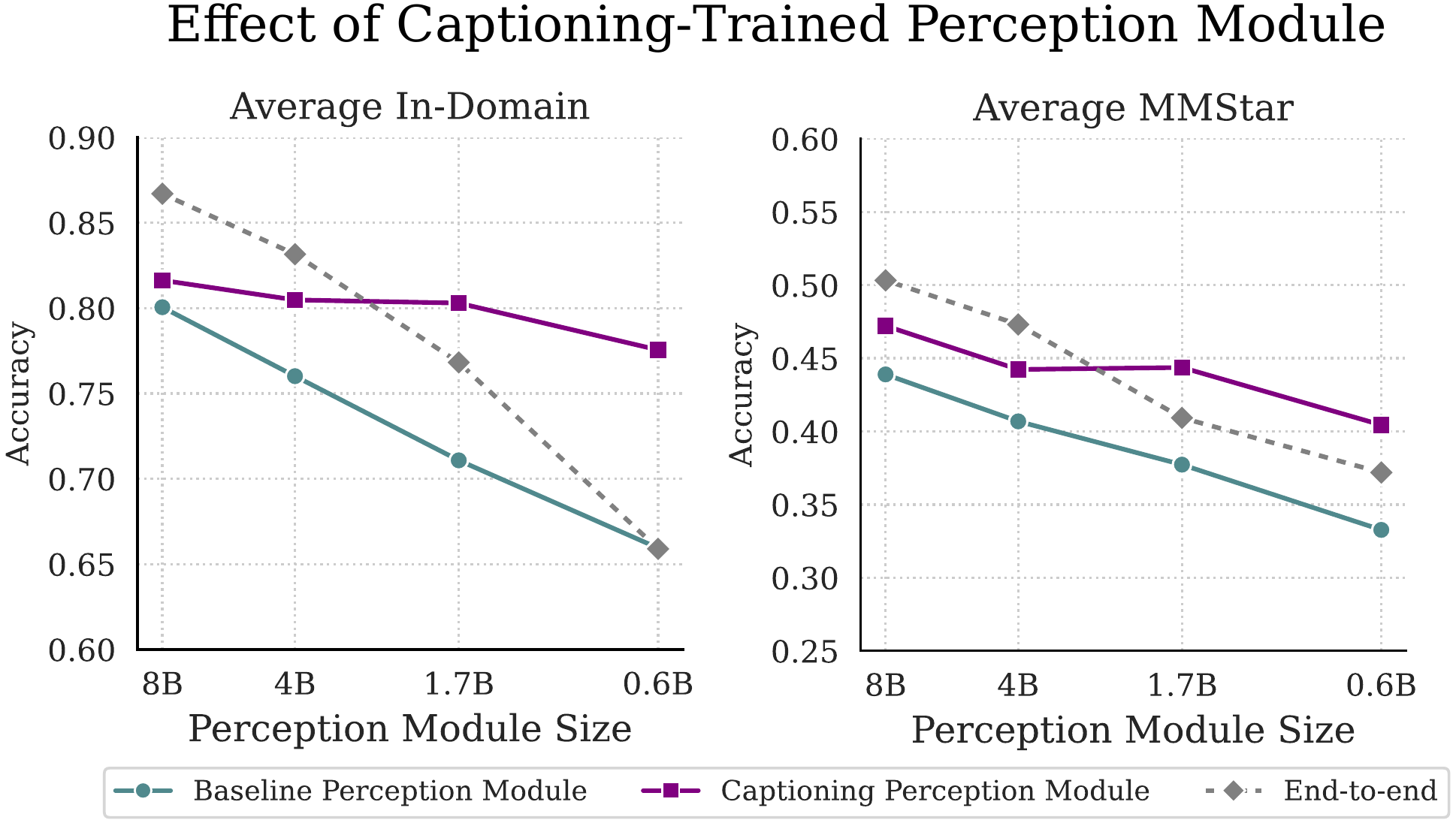}
    \caption{\textbf{Captioning alleviates perception bottleneck.} Decoupled frameworks use an 8B reasoning module.}
   \label{fig:captioning}
\end{figure}

\begin{figure}[ht]
    \centering

    \begin{subfigure}[b]{1.0\linewidth}
    \centering
        \includegraphics[width=1\linewidth]{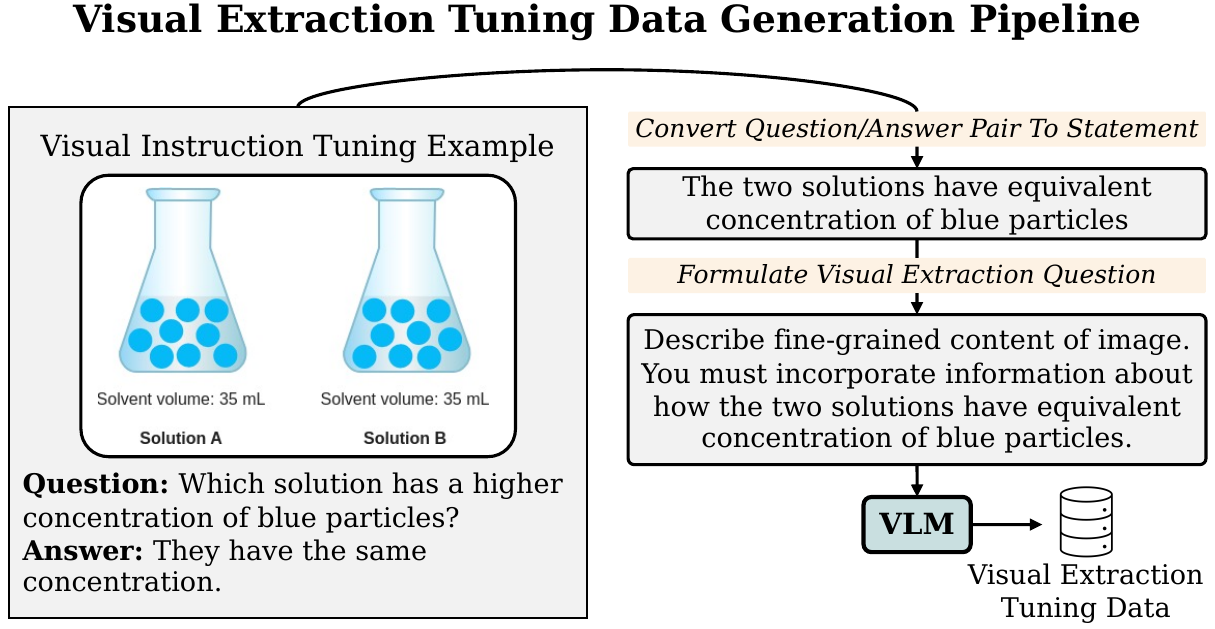}
    \end{subfigure}
    \\[1em]
    \begin{subfigure}[b]{1.0\linewidth}
    \centering
    \renewcommand{\arraystretch}{0.9} 
\setlength{\aboverulesep}{1pt}
\setlength{\belowrulesep}{1pt}
        \begin{tabular}{lccc}
            \toprule
            Perception Module & Size & In-domain & MMStar \\
            \midrule
            Captioning & 0.6B & 77.6 & 40.4\\
            \colorcell + Visual Extraction & \colorcell 0.6B & \colorcell 82.8 & \colorcell 44.0 \\
             \small \textcolor{mygreen}{$\Delta$} & & \small \bftable \textcolor{mygreen}{+5.2} & \small \bftable \textcolor{mygreen}{+3.6} \\
            \midrule
            Captioning & 1.7B & 80.3 & 44.4 \\
            \colorcell + Visual Extraction & \colorcell 1.7B & \colorcell  84.4 & \colorcell  49.0 \\
            \small \textcolor{mygreen}{$\Delta$} & & \small \bftable \textcolor{mygreen}{+4.1} & \small \bftable \textcolor{mygreen}{+4.6} \\
            \bottomrule
        \end{tabular}
    \end{subfigure}
    
    \caption{\textbf{Visual extraction tuning.} \textit{(Top)} \textbf{Simple pipeline for generating visual extraction tuning data.} Given a visual instruction tuning example, it is converted to a \textit{visual extraction} task by prompting a VLM to describe fine-grained visual details relevant to the original question. \textit{(Bottom)} \textbf{Visual extraction tuning enhances perception.} Post-training on visual extraction data improves both in-domain and out-of-domain (MMStar) performance. Size indicates the number of parameters of the perception module's LLM. All setups use an 8B reasoning module.
     }
   \label{fig:visual_extraction_tuning}
\end{figure}

\textbf{Captioning baseline.}
A simple way to unify perceptual skills for visual question answering is to train the perception module as a captioner. We therefore post-train the perception module on ALLaVA-4V \cite{chen2024allava}, a 950K caption dataset. As shown in Figure~\ref{fig:captioning}, this approach mitigates the effect of LLM downscaling and even outperforms end-to-end baselines at smaller scales (0.6B, 1.7B). However, \textbf{\textit{captioning introduces two key limitations}}. First, the two-stage framework is not merely captioning plus reasoning; the first stage should extract question-relevant visual details, which captioning does not teach. Second, visual instruction tuning often involves specialized, domain-specific data. Training solely on general captioning datasets limits domain-specific understanding, as the model is not taught to interpret specialized visual concepts present in those domains. Thus, an alternative approach is required to address these limitations.

\textbf{Visual extraction tuning.}
Here, we propose \textit{visual extraction tuning} as a solution to unify the perceptual abilities of the perception module while enabling it to extract question-relevant information and operate effectively across the diverse domains present in visual instruction tuning. Provided visual instruction data, we design a simple pipeline that transfers this data to the task of \textit{visual extraction}, where the goal is to generate all visual information relevant to answering the instruction, aligning precisely with the role of the perception module in the two-stage framework.

As shown in Figure~\ref{fig:visual_extraction_tuning}(Top), given a visual instruction tuning example, we first convert the question–answer pair into a declarative statement by prompting a model. We then integrate this declarative statement into a prompt that asks the model to describe fine-grained visual details, with explicit emphasis on information relevant to the declarative statement. Finally, this instruction, together with the image, is provided to a model to generate the visual extraction response. For simplicity, we use Qwen3VL-8B \cite{qwen3vl} throughout the entire process; although the first step is text-only, this model has shown strong performance on purely textual tasks. We apply this pipeline to 382K training samples corresponding to the assessed in-domain tasks, and post-train our captioning perception module with this data. Additional details on the generation process, including prompt templates and data examples, are provided in the supplemental material.

\textbf{Results.} As shown in Figure~\ref{fig:visual_extraction_tuning}(Bottom), we find that additionally post-training under the visual extraction tuning paradigm offers large performance improvements over the captioning baseline on both in-domain data and the out-of-domain MMStar benchmark. Specifically, in-domain performance increases by 5.2 when the perception module uses a 0.6B LLM and by 4.1 when it uses a 1.7B LLM. On the MMStar benchmark, performance improves by 3.6 for the 0.6B LLM and by 4.6 for the 1.7B LLM.

\takeaway{\textbf{Takeaway 3:} Visual extraction tuning proves an effective and efficient solution for alleviating the perception bottleneck of small multimodal models.}

\subsection{Step-by-step visual reasoning}
\label{sec:improving_reasoning}

Chain-of-Thought (CoT) reasoning is a widely studied method for improving LLM reasoning  \cite{wei2022chain, kojima2022large}. In our two-stage framework, although the reasoning module is not trained on visual data, text serves as an interface connecting perception and reasoning. Therefore, we expect that encouraging step-by-step in the reasoning module will directly enhance visual reasoning without requiring training.

\textbf{Approach.} The Qwen3 model \cite{yang2025qwen3}, which we utilize for the reasoning module, is capable of complex, multi-step reasoning by enabling thinking mode. Thus, we activate thinking mode and modify the prompt: instead of directly requesting the answer like before, we instruct the model to reason step-by-step. Since Qwen3 produces long reasoning chains, to improve efficiency we limit self-reflection with \textsc{NoWait} \cite{wang2025wait} and limit the thinking budget to 4096 tokens using \cite{Mueller2025LimitingQwen3Thinking}. Additional information is available in the supplementary material.

\begin{figure}[t]
    \centering
    \includegraphics[width=1\linewidth]{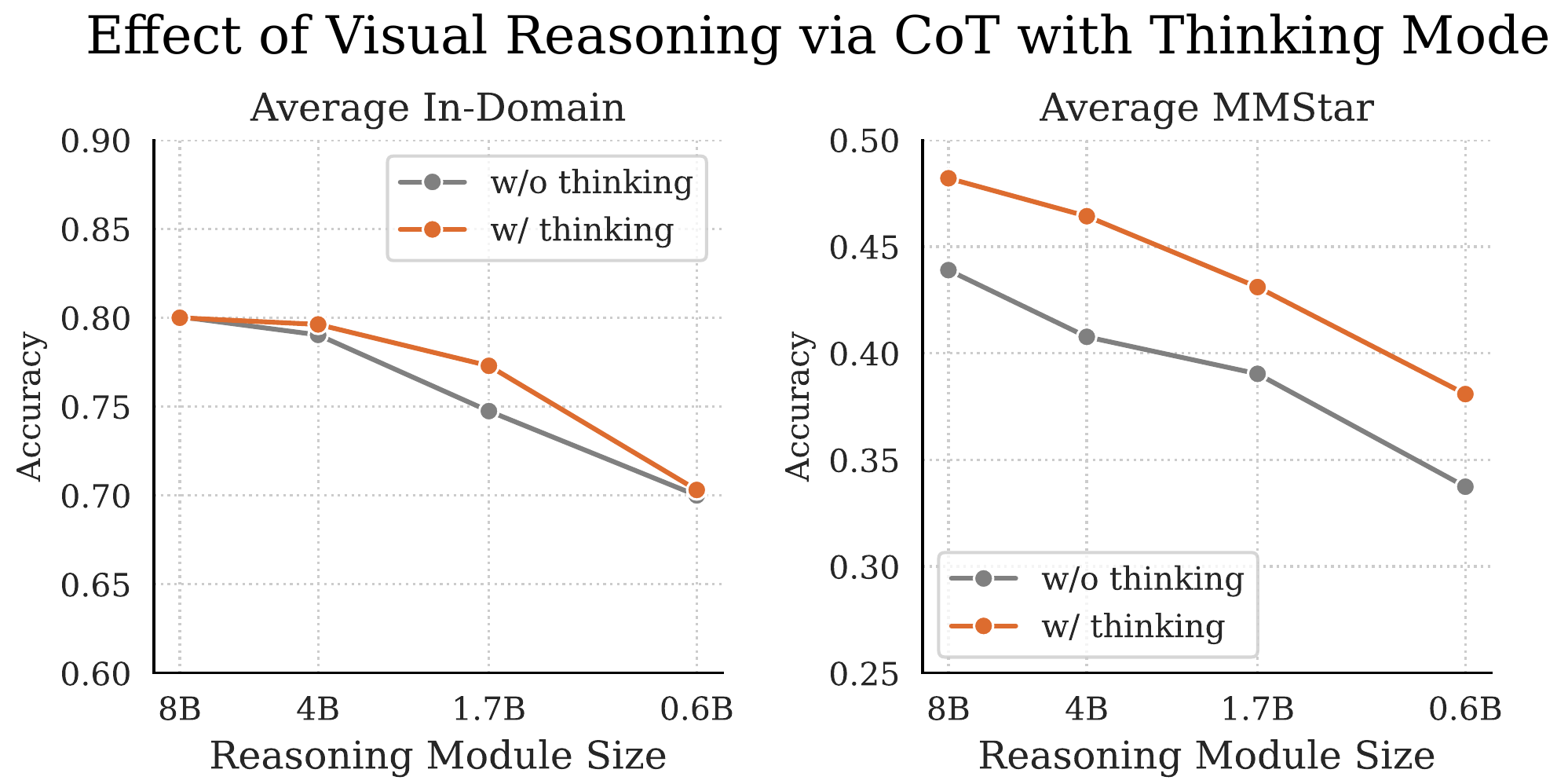}
    \caption{\textbf{CoT reasoning enhances in-domain and out-of-domain performance.} Performance gains exhibited at intermediate model scales (4B and 1.7B) for in-domain tasks, while out-of-domain performance improves across all LLM sizes. Both setups use 8B baseline perception module. Per-task plots provided in supplemental material.
    }
   \label{fig:thinking_results}
\end{figure}

\textbf{Results.} As shown in Figure~\ref{fig:thinking_results}, incorporating CoT reasoning substantially improves out-of-domain performance across all LLM sizes. For in-domain tasks, we observe a more nuanced behavior where the performance degradation under LLM downscaling becomes more concave when reasoning is enabled: the 8B and 0.6B models perform similarly with or without CoT, but at intermediate scales (4B and 1.7B), CoT yields notable gains. This suggests that \textbf{\textit{while CoT does not fully resolve the reasoning bottleneck in smaller multimodal models, it meaningfully enhances performance}}--particularly at mid-range LLM sizes, where it brings results closer to those of larger models.

\takeaway{\textbf{Takeaway 4:} Utilizing CoT boosts visual reasoning capabilities without requiring any supervision on visual data.}

\begin{table*}[t]
  \centering
   \begin{small}
  \begin{tabular}
  {lc@{\colpad}c@{\colpad}c@{\colpad}|@{\colpad}c@{\colpad}c@{\colpad}c@{\colpad}c@{\colpad}c@{\colpad}c@{\colpad}>{\columncolor{tablegray}}c@{\colpad}|@{\colpad}c@{\colpad}c@{\colpad}c@{\colpad}c@{\colpad}c@{\colpad}c@{\colpad}>{\columncolor{tablegray}}c@{\colpad}}
  &  & & & \multicolumn{7}{c}{In-Domain (Multiple-Choice \cite{zhang2025autoconverter})} & \multicolumn{7}{c}{Out-of-Domain (MMStar)}\\
  & \multicolumn{2}{c}{LLM Size}  & \rotatecol{\#Vis. Data} & \rotatecol{OCR-VQA} & \rotatecol{TextVQA} & \rotatecol{ScienceQA} & \rotatecol{VQAv2} & \rotatecol{GQA} & \rotatecol{VizWiz} & \rotatecol{Average} & \rotatecol{CP} & \rotatecol{FP} & \rotatecol{IR} & \rotatecol{LR} & \rotatecol{ST} & \rotatecol{Math} & \rotatecol{Average} \\
\midrule
\textit{End-to-End} &  & & & & & & & & & & & & & & & \\
  LLaVA-OneVision \cite{li2024llavaonevision} & \multicolumn{2}{c}{0.5B} & 8.8M & 69.5 & 77.2 & 55.7 & 75.7 & 73.6 & 74.7 & 71.1 & 63.2 & 31.1 & 42.1 & 35.8 & 30.0 & 31.4 & 39.0 \\
  InternVL2.5 \cite{chen2024internvl2} & \multicolumn{2}{c}{0.5B} & 64M & 79.8 & \underline{89.1} & \bftable 89.8 & 82.0 & \underline{75.4} & 83.0 & \underline{83.2} & \bftable 69.9 & 38.8 & 53.9 & 37.7 & \bftable 39.3 & 49.7 & 48.2 \\
 SmolVLM \cite{marafioti2025smolvlm} & \multicolumn{2}{c}{1.7B} & unk. & 72.9 & 81.4 & 79.7 & 75.5 & 70.6 & 75.1 & 75.9 & \underline{69.2} & 30.6 & 45.9 & 37.9 & 29.8 & 34.2 & 41.3 \\
Baseline (from \S\ref{sec:preliminary_exploration}) & \multicolumn{2}{c}{0.6B} & 1.0M & 41.1 & 71.3 & 67.9 & 71.2 & 69.5 & 74.5 &  65.9 & 58.1 & 30.4 & 39.3 & 35.1 & 27.4 & 32.9 & 37.2 \\
  Baseline (from \S\ref{sec:preliminary_exploration}) & \multicolumn{2}{c}{1.7B} & 1.0M & 73.4 & 83.4 & \underline{76.2} & 77.8 & 74.3 & 75.8 & 76.8 & 63.9 & 35.1 & 45.6 & 38.5 & 27.5 & 34.9 & 40.9\\
  \midrule
   \textit{Decoupled Models} &  P &  R & &  & & & & & & &  & & & & &  & \\
  PrismCaptioner \cite{qiao2024prism} & 1.8B & 70B & 1.9M & 89.2 & 72.7 & 64.6 & 77.8 & 66.0 & 82.3 & 75.4 & 64.0 & 38.8 & 55.8 & 36.7 & 23.0 & 33.1 & 41.9  \\
  PrismCaptioner \cite{qiao2024prism} & 7.0B & 70B & 1.9M & \underline{91.5} & 77.0 & 68.1 & 79.9 & 67.5 & 85.8 & 78.3 & 66.7 & 38.5 & \bftable 61.5 & 39.8 & 26.7 & 40.4 & 45.7 \\
  Baseline (from \S\ref{sec:who_blinks_first}) & 0.6B & 4.0B & 1.0M & 71.8 & 50.7 & 63.0 & 67.6 & 62.3 & 72.3 &  64.6 & 58.2 & 25.4 & 38.7 & 26.5 & 20.7 & 34.2  & 34.0\\
  Baseline (from \S\ref{sec:who_blinks_first}) & 1.7B & 4.0B & 1.0M & 79.4 & 59.4 & 65.0 & 71.6 & 64.5 & 76.4 & 69.4 & 62.2 & 30.4 & 46.3 & 32.0 & 29.2 & 35.9 & 39.4 \\
  \textsc{Caption+Think} & 0.6B & 1.7B & 2.0M & 84.9 & 80.6 & 60.6 & 74.7 & 66.2 & 83.0 & 75.0 & 60.7 & 37.2 & 51.9 & 38.9 & 27.0 & 42.4 & 43.0 \\
  \textsc{Caption+Think} & 1.7B & 4.0B & 2.0M & 89.2 & 84.8 & 68.9 & 80.5 & 72.1 & 84.3 & 80.0 & 64.6 & 37.6 & 53.4 & \underline{48.6} & 33.9 & \bftable 56.2 & \underline{49.0} \\
  \colorcell \textsc{Extract+Think}$^{\dagger}$ & \colorcell 0.6B & \colorcell 1.7B & \colorcell 0.4M & \colorcell 86.9 & \colorcell 79.8 & \colorcell 69.9 & \colorcell 76.6 & \colorcell 72.5 & \colorcell 82.1 & \colorcell 78.0 & \colorcell 65.2 & \underline{\colorcell 41.7} & \colorcell 49.7 & \colorcell 37.5 & \colorcell 21.9 & \colorcell 39.8 & \colorcell 42.6  \\
  \colorcell \textsc{Extract+Think}$^{\dagger}$ & \colorcell 1.7B & \colorcell 4.0B & \colorcell 0.4M & \colorcell \underline{\colorcell 91.5} & \colorcell 84.0 & \colorcell 71.3 & \colorcell \bftable 84.6 & \colorcell \bftable 77.8 &  \underline{\colorcell 86.9} & \colorcell 82.7 & \colorcell 64.4 & \colorcell 40.7 & \colorcell 58.4 & \colorcell 46.3 & \underline{\colorcell 35.5} & \colorcell 43.1 & \colorcell 48.1 \\
   \colorcell \textsc{Extract+Think} & \colorcell 0.6B & \colorcell 1.7B & \colorcell 2.4M & \colorcell 89.4 & \colorcell 81.8 & \colorcell 72.2 & \colorcell 78.0 & \colorcell 74.7 & \colorcell 85.6 & \colorcell 80.3 & \colorcell 64.5 & \underline{\colorcell 41.7} & \colorcell 54.9 & \colorcell 43.0 & \colorcell 28.3 & \colorcell 47.3 & \colorcell 46.6 \\
  \colorcell \textsc{Extract+Think} & \colorcell 1.7B & \colorcell 4.0B & \colorcell 2.4M & \colorcell \bftable 92.9 & \colorcell \bftable 90.1 & \colorcell 75.2 & \underline{\colorcell 84.4} & \colorcell \bftable 77.8 & \colorcell \bftable 91.3 & \colorcell \bftable 85.3 & \colorcell 68.5 & \colorcell \bftable 47.8 & \underline{\colorcell 59.2} & \colorcell \bftable 53.3 & \colorcell 33.0 & \underline{\colorcell 53.8} & \colorcell \bftable 52.6\\

  \end{tabular}
    \end{small}
\caption{\textbf{\textsc{Extract+Think} demonstrates extreme effectiveness as a generalist small multimodal model.}
Even the smaller \textsc{Extract+Think} variant surpasses LLaVA-OneVision-0.5B by up to 19.5\% while using 73\% fewer visual samples, and outperforms the larger PrismCaptioner model on both in-domain and out-of-domain tasks with a perception module roughly 12$\boldsymbol{\times}$ smaller and a reasoning module 41$\boldsymbol{\times}$ smaller. The \textsc{Extract+Think}$^{\dagger}$ configuration, trained from scratch under the visual extraction tuning paradigm, demonstrates robust performance using very minimal data. \#Vis. Data denotes the amount of visual data used for training (excluding the connector pre-training stage). P=Perception Module, R=Reasoning Module. For MMStar, CP=Coarse Perception, FP=Fine-grained Perception, IR=Instance Reasoning, LR=Logical Reasoning, ST=Science \& Technology. The best results are bolded and the second best are underlined. Inference latency analysis included in supplementary material.}
  \label{tab:final_result_table}
\end{table*}

\subsection{Distilling insights}
\label{sec:final_results}

Guided by these insights, we now present our final approach, \textsc{Extract+Think}. Specifically, we employ the perception module trained under our proposed visual extraction paradigm from \S\ref{sec:improving_perception} and a reasoning module enhanced with CoT reasoning from \S\ref{sec:improving_reasoning}. Based on our finding that CoT does not fully resolve the reasoning bottleneck in smaller multimodal models, we adopt a larger LLM for the reasoning module than for the perception module (while keeping both models within a lightweight regime). We present two configurations: one where the perception module’s LLM size is 0.6B and the reasoning module’s is 1.7B, and a larger setup with 1.7B and 4B, respectively.

For the perception module, we test two configurations--one post-trained under the visual extraction paradigm starting from a captioning model (as is done in \S\ref{sec:improving_perception}), and another trained from scratch without prior instruction tuning or captioning. We compare \textsc{Extract+Think} against both end-to-end baselines and other decoupled methods, including PrismCaptioner~\cite{qiao2024prism}, the original decoupled setup from \S\ref{sec:who_blinks_first}, and the captioning baseline in \S\ref{sec:improving_perception} (denoted \textsc{Caption+Think}).

\textbf{Results.} \textbf{\textsc{Extract+Think} \textit{substantially outperforms decoupled baselines and even competes with end-to-end models trained at vast scale}}. As shown in Table~\ref{tab:final_result_table}, even our smaller variant surpasses the largest PrismCaptioner model on both in-domain and out-of-domain tasks, with a perception module LLM roughly \textbf{\textit{12$\boldsymbol{\times}$ smaller}} and a reasoning module \textbf{\textit{41$\boldsymbol{\times}$ smaller}}. It also outperforms LLaVA-OneVision-0.5B by 12.9\% on in-domain data and 19.5\% on the out-of-domain MMStar benchmark, while using \textbf{\textit{73\%}} fewer visual samples.

\textbf{\textit{Visual extraction tuning offers a data-efficient solution for generalist small multimodal models}}. Looking at our configuration trained from scratch without prior visual training (denoted as \textsc{Extract+Think}$^{\dagger}$ in Table~\ref{tab:final_result_table}), the smaller variant improves over LLaVA-OneVision-0.5B by 9.7\% on in-domain data while using \textbf{\textit{95\%}} fewer visual samples. This setup also outperforms the 1.7B baseline trained directly on the in-domain instruction tuning data, and even exceeds the in-domain performance of the comparable \textsc{Caption+Think} configuration, which was trained on both the in-domain instruction tuning data and 950K additional captioning examples. Overall, these results demonstrate that visual extraction tuning is an extremely effective and efficient paradigm for training small multimodal models.

%% file: sec/5_conclusion.tex
\section{Conclusion}

In this work, we provide a systematic study of how language model downscaling affects multimodal task performance, revealing that visually demanding tasks are disproportionately impacted. Through a decoupled analysis, we identify that both foundational perception and downstream reasoning abilities are central bottlenecks when downscaling LLMs. To address these limitations, we introduce a two-stage perception–reasoning framework that employs visual extraction tuning to enhance the model’s ability to extract relevant visual details across tasks and applies step-by-step reasoning over the extracted data without requiring additional visual training. Our final approach establishes a highly parameter- and data-efficient paradigm for training small multimodal models, setting a new standard for efficiency and performance in this space.

This work lays the groundwork for future research on downscaling of multimodal models. On the analysis side, future studies can explore downscaling across a broader range of model sizes, assess how the downscaling of visual representations compares to that of language models, and incorporate data size as a variable to examine how downscaling behavior varies across different scales. On the methodological side, future research can further investigate the visual extraction tuning paradigm in comparison to visual instruction tuning and evaluate its effectiveness with larger language models.

\clearpage

%% file: sec/6_acknowledgements.tex
\noindent \textbf{Acknowledgments.} This work is supported in part by the National Science Foundation (NSF) under Grant No. 2026498 and the NSF Graduate Research Fellowship Program under Grant No. DGE-2146755 (for M.E.). Any opinions, findings, and conclusions or recommendations expressed in this material are those of the authors and do not necessarily reflect the views of any other entity.

%% file: sec/X_suppl.tex
\clearpage
\maketitlesupplementary

\renewcommand{\thesection}{A\arabic{section}}
\renewcommand{\thefigure}{A\arabic{figure}}
\renewcommand{\thetable}{A\arabic{table}}

\setcounter{section}{0}
\setcounter{figure}{0}
\setcounter{table}{0}

\section{Additional LLM Downscaling Results and Details}

\begin{figure*}[b]
    \centering
    \includegraphics[width=1\linewidth]{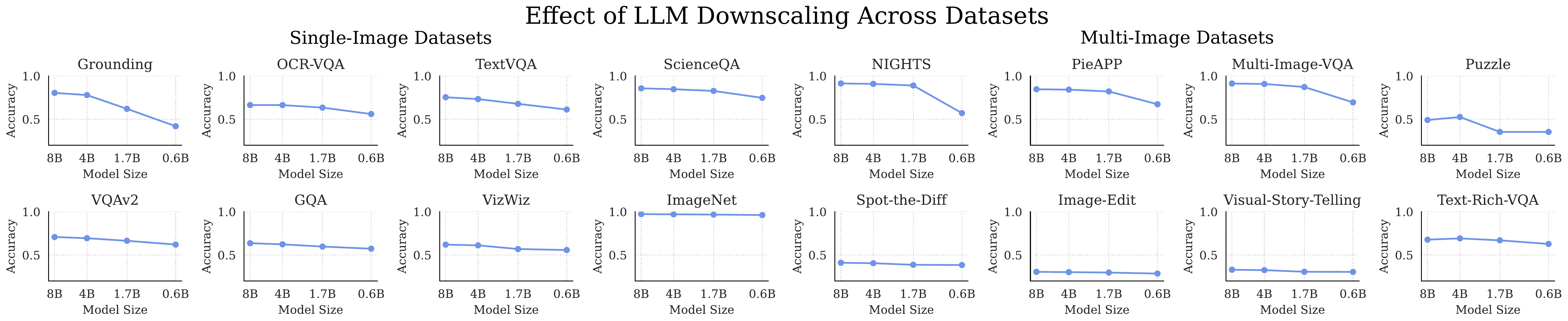}
    \caption{Performance dropoff from downscaling LLM across all datasets.
    }
   \label{fig:supp_per_task_results}
\end{figure*}
\textbf{Per-tasks results}. We present plots showing the performance dropoff from LLM downscaling across all evaluated tasks in Figure~\ref{fig:supp_per_task_results}. As described in the main text, most tasks exhibit minimal performance decline when downscaling the LLM, except for a handful of vision-centric tasks that exhibit substantially larger drops (e.g., Grounding, NIGHTS, PieAPP).

\begin{figure*}[b]
    \centering
    \includegraphics[width=1\linewidth]{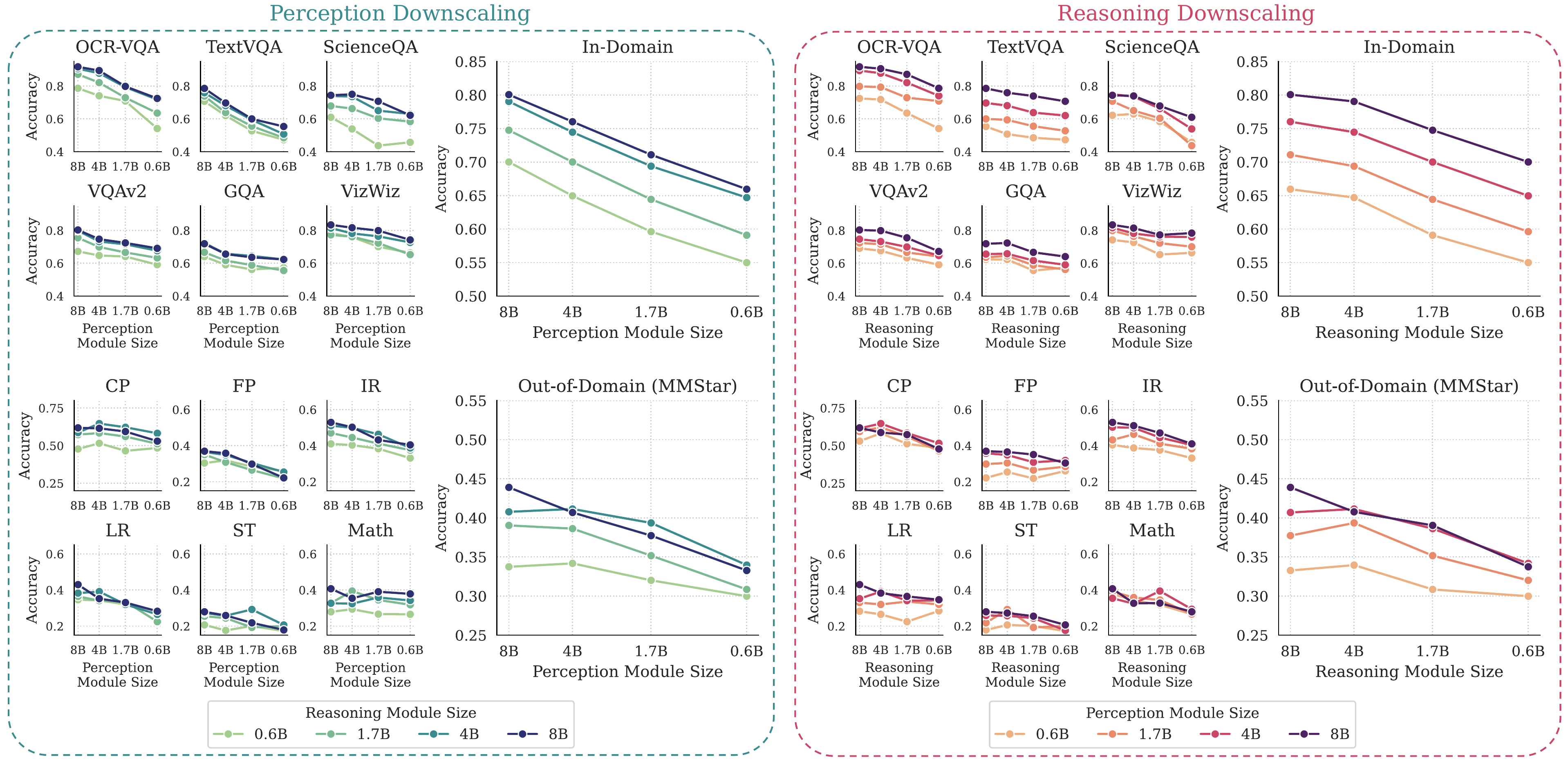}
    \caption{Full decoupled results. CP=Coarse Perception, FP=Fine-grained Perception, IR=Instance Reasoning, LR=Logical Reasoning, ST=Science \& Technology.
    }
   \label{fig:supp_decoupled_results}
\end{figure*}

\textbf{Full decoupled results}. We plot the performance dropoff from LLM downscaling of the perception and reasoning modules in Figure~\ref{fig:supp_decoupled_results}. We find that LLM downscaling of either module leads to performance degradation across a wide range of tasks. Notably, downscaling the perception module has a large effect on both tasks assessing perception (e.g., OCR-VQA, Fine-grained Perception) and reasoning (e.g., Logical Reasoning). One exception is Math, where LLM downscaling of the perception module has little impact. We expect this is because mathematical ability is limited primarily by the downstream process of operating on visual information (reasoning) rather than by the foundational perception ability.

\begin{figure*}[b]
    \centering
    \includegraphics[width=.83\linewidth]{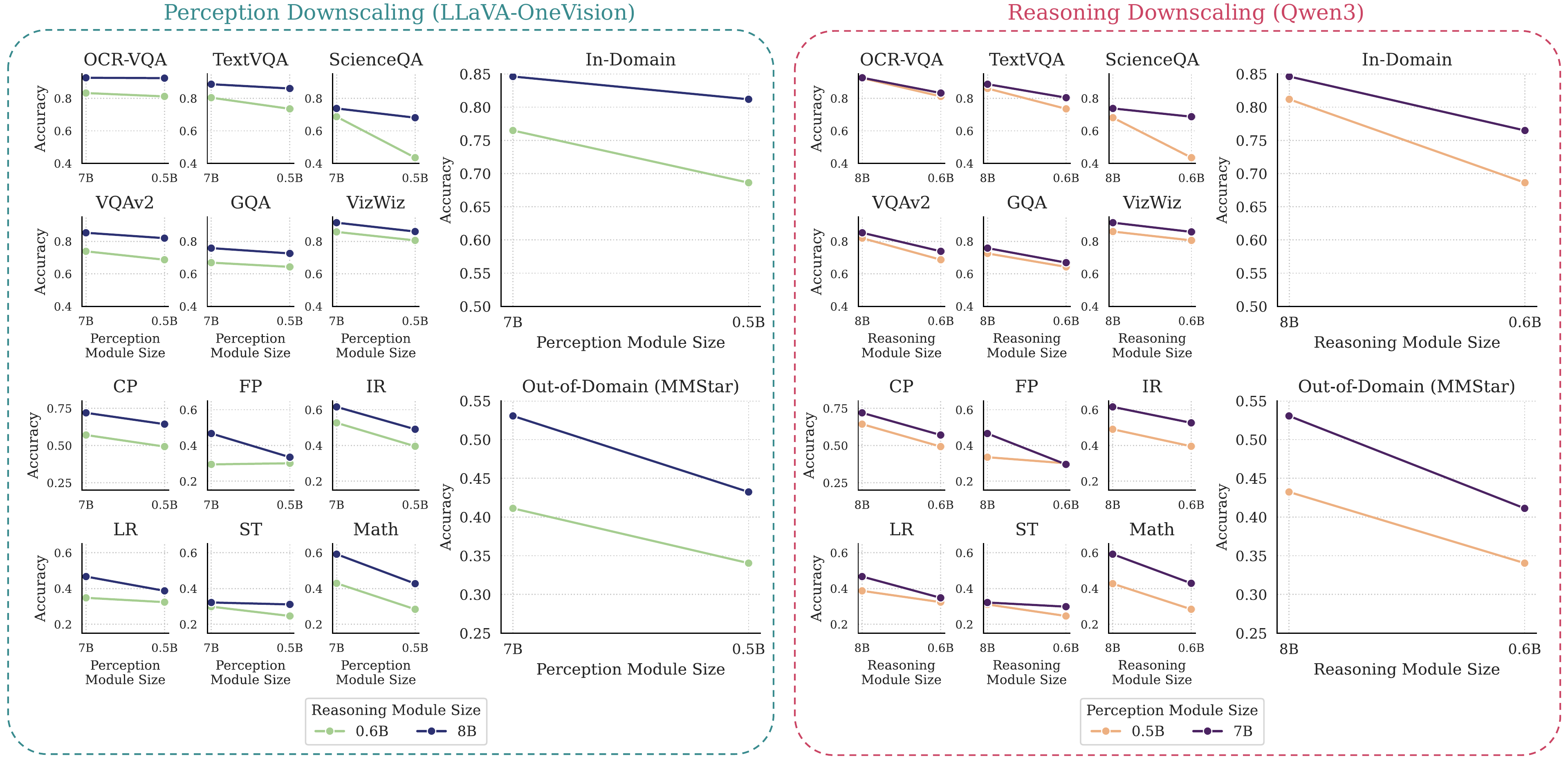}
    \caption{Decoupled analysis using LLaVA-OneVision as the perception module and Qwen3 as the reasoning module. CP=Coarse Perception, FP=Fine-grained Perception, IR=Instance Reasoning, LR=Logical Reasoning, ST=Science \& Technology.
    }
   \label{fig:supp_decoupled_llava_ov_qwen3}
\end{figure*}

\newpage
\textbf{LLaVA-OneVision as the perception module}. While our main analysis used models trained from scratch for a controlled study, here we also experiment with using LLaVA-OneVision ($\in{0.5\text{B}, 7\text{B}}$) as the perception module in the decoupled framework. We first present decoupled results using the same reasoning module as in our experiments (Qwen3). As shown in Figure~\ref{fig:supp_decoupled_llava_ov_qwen3}, this configuration produces results that are largely consistent with those obtained using our controlled model as the perception module, where LLM downscaling of either module hinders performance. We do observe, however, that downscaling the perception module has a smaller effect than in our controlled study for 

\begin{figure*}[b]
    \centering
    \includegraphics[width=.83\linewidth]{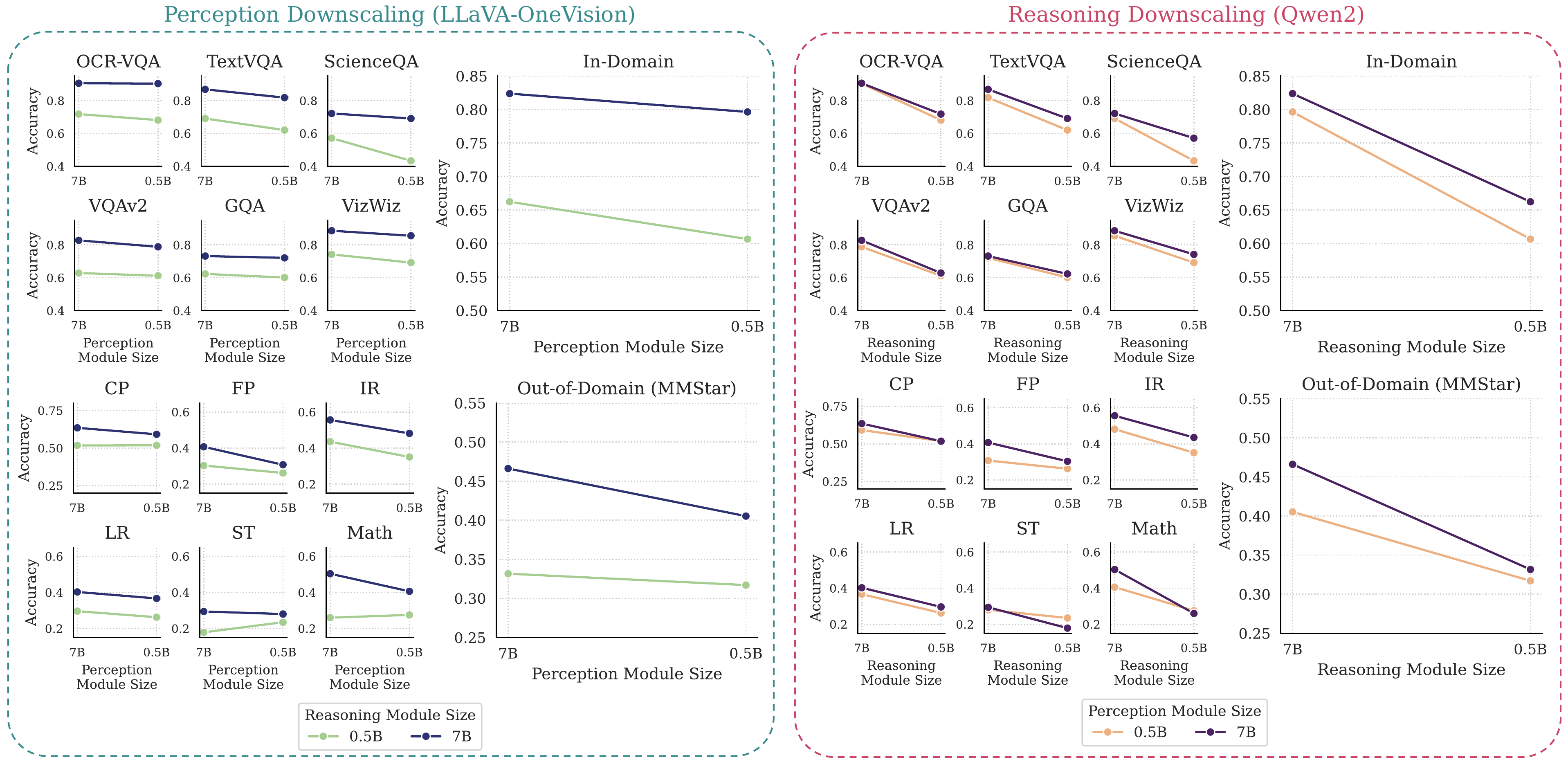}
    \caption{Decoupled analysis using LLaVA-OneVision as the perception module and Qwen2 as the reasoning module. CP=Coarse Perception, FP=Fine-grained Perception, IR=Instance Reasoning, LR=Logical Reasoning, ST=Science \& Technology.
    }
   \label{fig:supp_decoupled_llava_ov_qwen2}
\end{figure*}
\newpage \noindent in-domain data.
This is likely because LLaVA-OneVision includes extensive training on captioning, which we demonstrate alleviates the perception bottleneck.

Additionally, we experiment with using Qwen2 as the reasoning model in this setup (the LLM used in LLaVA-OneVision). As shown in Figure~\ref{fig:supp_decoupled_llava_ov_qwen2}, relative to the earlier results using Qwen3 as the reasoning module, we see a larger impact from downscaling the reasoning module on the in-domain tasks, and overall performance is weaker than when using Qwen3. This outcome is not surprising, as Qwen3 has demonstrated stronger performance than Qwen2 on textual tasks, particularly for smaller model variants.

\clearpage

\textbf{Prompts for decoupled analysis}. We list prompt templates for our decoupled perception and reasoning analysis in Figure~\ref{fig:prompt_decoupled_framework}. These prompts follow the Prism framework, except that the initial instruction for obtaining question-specific information is run offline using the same model throughout, ensuring that deriving question-specific instructions does not influence our analysis of perception and reasoning downscaling. Thus, the question-specific information inserted into the perception module prompt is consistent across all model setups.

\textbf{Perception bottleneck across model families.} We additionally analyze Gemma3 (12B $\rightarrow$ 4B) and InternVL2.5 (8B $\rightarrow$ 2B), which offer multiple LLM sizes with a fixed vision encoder, enabling a controlled analysis. As shown in

\begin{figure*}[b]
\centering
\begin{subfigure}{1.\linewidth}
    \centering
    \begin{promptboxcompact}{Question-specific Instruction Prompt}{black}\input{prompts/question_specific_instruction.tex}
    \end{promptboxcompact}
\end{subfigure}

\vspace{1.5em}
\begin{subfigure}{1.\linewidth}
    \centering
    \begin{promptboxcompact}{Perception Module Prompt}{perceptionblue}\input{prompts/perception_module_instruction}
    \end{promptboxcompact}
\end{subfigure}

\vspace{1.5em}
\begin{subfigure}{1.\linewidth}
    \centering
    \begin{promptboxcompact}{Reasoning Module Prompt (w/o thinking)}{reasoningpink}\input{prompts/reasoning_module_instruction}
    \end{promptboxcompact}
\end{subfigure}

\vspace{1.5em}
\begin{subfigure}{1.\linewidth}
    \centering
    \begin{promptboxcompact}{Reasoning Module Prompt (w/ thinking)}{reasoningpink}\input{prompts/reasoning_module_instruction_with_thinking}
    \end{promptboxcompact}
\end{subfigure}
\caption{Prompt templates for the decoupled perception / reasoning analysis.}
\label{fig:prompt_decoupled_framework}
\end{figure*}

\newpage 

\noindent Table~\ref{tab:generalization}, evaluating on MMStar using the same procedure as \S3.3, we observe consistent trends of decreasing performance when downscaling the perception module's LLM.

\begin{table}[h]
\centering
\begin{small}
\begin{tabular}{cccc}
\toprule
P Module & R Module & P LLM Size & Avg. Acc. Drop \\
\midrule
Ours (\S3) & Qwen3 & 8B $\rightarrow$ 1.7B & 3.30 \\
InternVL2.5 & InternLM2     & 8B $\rightarrow$ 2B   & 4.33 \\
Gemma3 & Gemma3 & 12B $\rightarrow$ 4B  & 5.26 \\
\bottomrule
\end{tabular}
\end{small}
\caption{Effect of downscaling the perception module on MMStar accuracy. For each model family, we vary the perception module's LM size while holding the reasoning module fixed, and report the accuracy drop averaged across both reasoning module sizes. P=Perception, R=Reasoning.}
\label{tab:generalization}
\end{table}

\clearpage

\section{Additional Visual Extraction Tuning Details}

\textbf{Visual extraction tuning data generation pipeline}. We present the prompt templates for our pipeline generating visual extraction tuning data in Figure~\ref{fig:visual_extraction_tuning_prompts}. In the first stage, each question–answer pair in a visual instruction example is converted into a declarative statement. These statement(s) are then inserted into a prompt that instructs the model to describe fine-grained visual details, with explicit emphasis on the information contained in the visual instruction example.

\begin{figure*}[b]
\centering
\begin{subfigure}{1.\linewidth}
    \centering
    \begin{promptbox}{Question-answer Pair to Declarative Statement Prompt}{black}\input{prompts/qa_to_declarative_statement}
    \end{promptbox}
\end{subfigure}

\vspace{2em}
\begin{subfigure}{1.\linewidth}
    \centering
    \begin{promptbox}{Visual Extraction Response Generation Prompt}{black}\input{prompts/visual_extraction_tuning_generation_instruction}
    \end{promptbox}
\end{subfigure}

\caption{Prompts used for visual extraction tuning data generation pipeline.}
\label{fig:visual_extraction_tuning_prompts}
\vspace{19em}
\end{figure*}

\begin{figure*}[t]
    \centering
    \includegraphics[width=.86\linewidth]{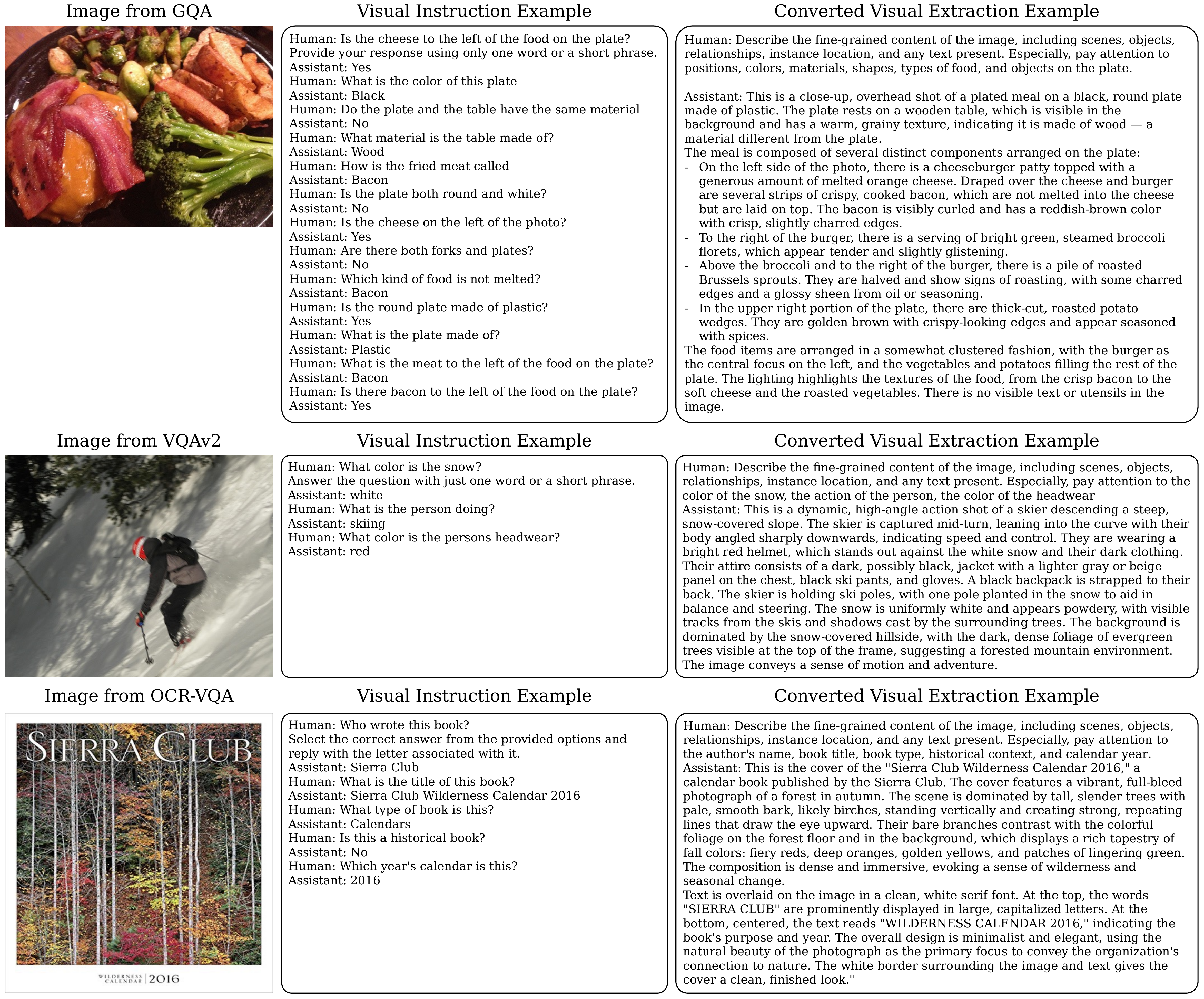}
    \caption{Visual extraction tuning data examples.
    }
   \label{fig:visual_extraction_tuning_examples}
\end{figure*}

\newpage \textbf{Visual extraction tuning data examples}. We provide examples produced by our visual extraction tuning data generation pipeline in Figure~\ref{fig:visual_extraction_tuning_examples}. We see that the generated examples preserve the original information required to answer the questions from the visual instruction while also offering a more detailed and comprehensive depiction of the specific image content.

\clearpage

\section{Additional Step-by-step Reasoning Details and Results}

\textbf{NoWait Setup.} To reduce overthinking of Qwen3 with thinking mode enabled, we use a logits processor that suppresses self-reflection tokens. Namely, we mask the logits of any token that contains one of the following keywords: \{wait, alternatively, hmm, but, however, alternative, another, check, double-check, oh, maybe, verify, other,

\begin{figure*}[b]
    \centering
    \includegraphics[width=.85\linewidth]{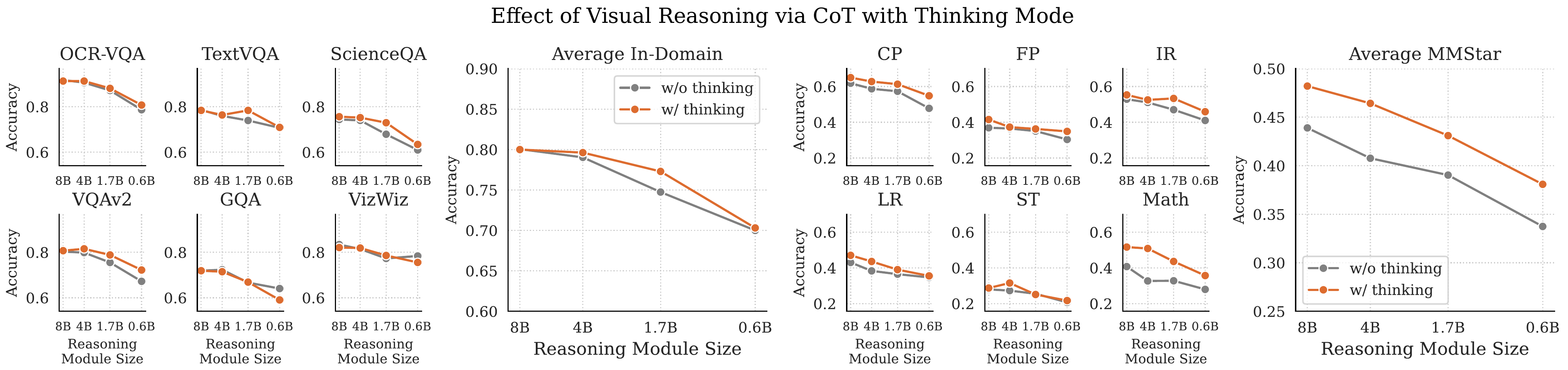}
    \caption{Full results showing impact of step-by-step reasoning on in-domain and out-of-domain (MMStar) performance.
    }
   \label{fig:full_thinking_results}
\end{figure*}

 \newpage \noindent again, now, ah, anyway, anyhow\}, while manually excluding words that only contain a keyword as a substring but are not reflexive (e.g., waiter).
 
 \textbf{Full results.} We present results from performing step-by-step visual reasoning across all tasks in Figure~\ref{fig:full_thinking_results}. Expectedly, we find that Math heavily benefits from CoT reasoning (consistent with findings in text-only Math tasks).

 \section{Additional \textsc{Extract+Think} Analyses}

\begin{figure}[h]
    \centering
    \includegraphics[width=1\linewidth]{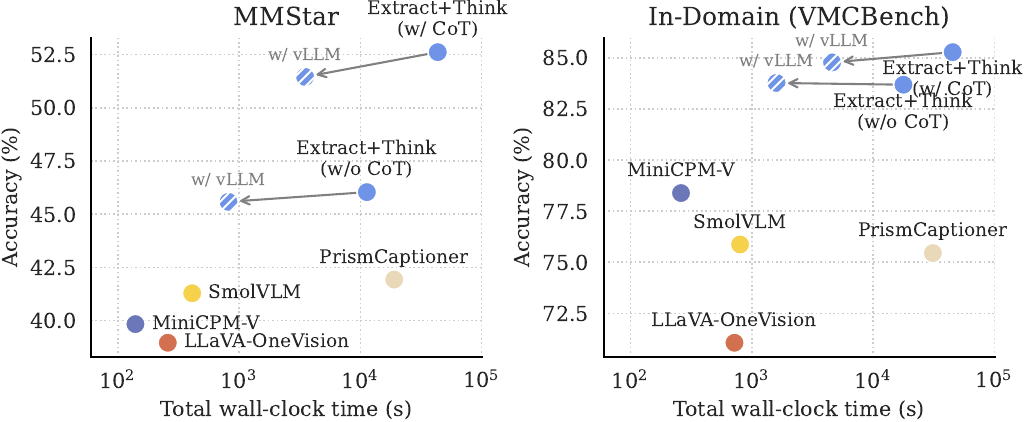}
    \caption{Inference latency on 1 L40s using Transformers (BS=1) or vLLM (BS=32), evaluated on MMStar and In-Domain (VMCBench).}
   \label{fig:inference_latency}
\end{figure}

 \textbf{Inference latency analysis.} In Figure~\ref{fig:inference_latency}, we characterize the tradeoff between our approach's parameter/data efficiency and its inference latency. We plot latency across end-to-end baselines, PrismCaptioner (1.8/70B), and our \textsc{Extract+Think} method (1.7/4B w/ and w/o CoT). While end-to-end approaches incur lower latency from generating far fewer tokens, our method achieves 27.4\% improvement on MMStar and 8.8\% on VMCBench tasks. Additionally, vLLM greatly increases generation throughput for longer output sequences, narrowing the gap, particularly at a higher batch size.

Comparing against PrismCaptioner underscores the benefit of visual extraction tuning: even w/o CoT, our method improves performance by 9.80\% on MMStar and 10.9\% on VMCBench while using a smaller reasoning module and reducing latency. This indicates that increasing the token budget via a two-stage design alone does not drive performance; rather, it is the visual extraction tuning that drives gains within this framework. Applying test-time scaling through CoT further improves performance, particularly on MMStar.

\begin{figure}[t]
    \centering
    \includegraphics[width=1\linewidth]{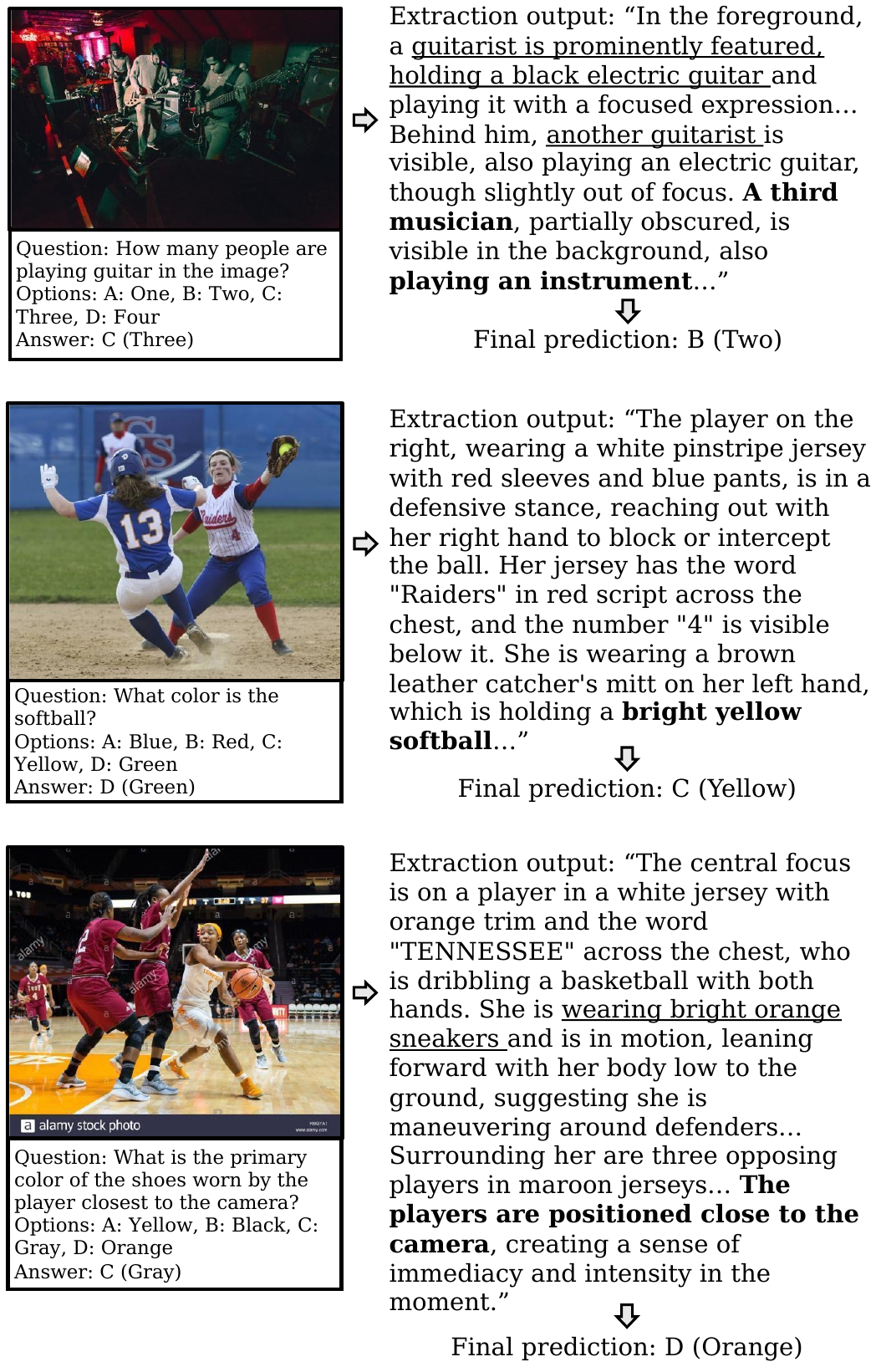}
    \caption{Examples showing failure modes of two-stage framework for visual question answering (using \textsc{Extract+Think} w/ 1.7B perception module).}
   \label{fig:information_bottleneck}
\end{figure}

\textbf{Information bottleneck in two-stage design.}  
While the two-stage approach offers a modular framework for independently studying and improving perception and reasoning and has been shown to enable strong performance, a drawback is that its design requires all relevant visual information for answering the question to be captured and adequately described in the initial stage. This stage acts as an intermediate representation: the reasoning module bases its final response on this description rather than the image itself. In Figure~\ref{fig:information_bottleneck}, we show examples where this leads to loss of critical visual information, ultimately leading the reasoning module astray. For instance, a musician may be described as playing an instrument without specifying which one, or players may be described without indicating their exact, relative spatial relationship to the camera, rendering certain questions unanswerable. Nonetheless, the proposed visual extraction tuning approach is specifically designed to mitigate such issues by targeting the extraction of question-relevant visual details and has been shown to effectively improve performance across evaluated tasks.

%% file: prompts/question_specific_instruction.tex
Your task is to give a concise instruction about what basic elements are needed to be described based on the given question. Ensure that your instructions do not cover the raw question, options or thought process of answering the question.

Examples:

Question: In which period the number of full time employees is the maximum?

Contents to observe: the number of full time employees

Question: What is the value of the smallest bar?

Contents to observe: the heights of all bars and their values

Question: What is the main subject of the image?

Contents to observe: the central theme or object

Question: What is the position of the catcher relative to the home plate?

Contents to observe: the spatial arrangement of the objects 

Question: What is the expected ratio of offspring with white spots to offspring with solid coloring? Choose the most likely ratio.

Contents to observe: the genetic information

Now, perform the task, and format your answer as "Contents to observe:"

Question: {\color{brown}<question>}

%% file: prompts/perception_module_instruction.tex
Describe the fine-grained content of the image, including scenes, objects, relationships, instance location, and any text present.

Especially, pay attention to <question-specific info>

%% file: prompts/reasoning_module_instruction.tex
You are an excellent text-based reasoning expert. You are required to answer the question based on the detailed description of the image.

Description: {\color{perceptionblue}<description>}

Question: {\color{brown}<question>}

Answer directly with the option\'s letter in the format of "Answer:". Do not add anything other than the letter answer after "Answer:".

%% file: prompts/reasoning_module_instruction_with_thinking.tex
You are an excellent text-based reasoning expert. You are required to answer the question based on the detailed description of the image.

Description: {\color{perceptionblue}<description>}

Question: {\color{brown}<question>}

Please reason step by step, and give the final answer on the last line by itself in the format of "Answer:". Do not add anything other than the letter answer after "Answer:".

%% file: prompts/qa_to_declarative_statement.tex
Your task is to convert each question–answer pair about an image into a concise, fully self-contained declarative statement. The resulting statements should be understandable on their own, without requiring the reader to refer to the original question.
\\
\\
\texttt{\% for each QA pair in conv include:}

\hspace{2em}Question: {\color{brown}<question>}

\hspace{2em}Answer: {\color{brown}<answer>}
\\
\\
\texttt{\% if len(conv) > 1 include:}

\hspace{2em}As there are <len(conv)> questions, you should respond with <len(conv)> statements. Include each statement on its own line

Declarative Statement(s):

%% file: prompts/visual_extraction_tuning_generation_instruction.tex
Your task is to describe the fine-grained content of the image, including scenes, objects, relationships, instance location, and any text present.

As part of your description, you should incorporate the following information about the image.

<declarative statements>

Description: